\documentclass[10pt]{article} 
\usepackage[preprint]{tmlr}
\usepackage{amsmath,amsfonts,bm}

\def\eqref#1{equation~\ref{#1}}
\def\1{\bm{1}}

\DeclareMathAlphabet{\mathsfit}{\encodingdefault}{\sfdefault}{m}{sl}
\SetMathAlphabet{\mathsfit}{bold}{\encodingdefault}{\sfdefault}{bx}{n}

\usepackage{hyperref}
\usepackage{url}
\usepackage{graphicx}
\usepackage{adjustbox}
\usepackage{amsmath}
\usepackage{textcomp}
\usepackage{graphicx}
\usepackage{amsfonts}
\usepackage{amssymb}
\usepackage{amsthm}  
\usepackage{subfig}
\usepackage{tabularx}
\usepackage{multirow}
\usepackage{makecell}
\usepackage{adjustbox}
\usepackage{url}
\usepackage{caption}

\title{Deep Discriminative Feature Fusion of Convolutional and Attention-Based Recurrent Representations for Subject-Independent ECG Stress Recognition}

\author{\name Zeeshan Ahmad \email zeeshan.ahmad@unfc.ca \\
      \addr Master of Computer Science and Data Analytics \\ University of Niagara Falls, Ontario, Canada
      \AND
      \name Muhammad Rehman Zafar \email muhammadrehman.zafar@torontomu.ca \\
      \addr Ted Rogers School of Information Technology Management \\ Toronto Metropolitan University, Ontario, Canada
}

\def\month{MM}  % Insert correct month for camera-ready version
\def\year{YYYY} % Insert correct year for camera-ready version
\def\openreview{\url{https://openreview.net/forum?id=XXXX}} % Insert correct link to OpenReview for camera-ready version

\renewcommand{\headrulewidth}{0pt}

\begin{document}

\maketitle

\begin{abstract}
Stress has become a significant concern because of its negative impact on both physical and mental health and therefore drives the need for reliable automated systems to assess stress levels. Electrocardiogram (ECG) signals provide a non-invasive and objective measure of physiological stress responses and have emerged as a promising modality for continuous stress monitoring. While deep learning techniques have achieved encouraging results in ECG-based stress detection, many studies focus only on  one type of feature representation, and therefore limit their ability to capture both the morphological patterns and temporal dynamics embedded in ECG signals. To address this limitation, in this paper, we propose three deep learning-based feature fusion frameworks for ECG-based stress classification. These are Residual Feature Fusion (RFF), Bidirectional Feature-Wise Cross-Attention Fusion (BFCAF), and a novel Cross-Attention Residual Fusion (CARF) framework that combines the strengths of residual learning and bidirectional cross-attention within a unified architecture. A 1D Convolutional Neural Network (1D-CNN) and a BiLSTM-Attention network are used to extract complementary morphological and temporal representations from ECG signals, respectively. The proposed fusion frameworks leverage these heterogeneous representations to improve feature interaction, information preservation, and stress discrimination. The proposed frameworks are evaluated using subject-independent Leave-One-Subject-Out Cross-Validation (LOSOCV) on the RML and WESAD datasets for three-class stress classification. Experimental results show that the proposed feature fusion methods consistently improve performance over individual CNN and BiLSTM models, with CARF achieving the highest accuracies of 83.1\% and 77.91\% on the RML and WESAD datasets, respectively.
\end{abstract}

\section{Introduction}
Stress is widely recognized as one of the major factors affecting both physical and mental health, contributing to cardiovascular disorders, anxiety, depression, reduced cognitive performance, and decreased quality of life. Since stress has become increasingly prevalent in modern society, therefore, the development of reliable stress assessment architectures has attracted considerable attention of the researchers in healthcare, affective computing, and wearable sensing technologies. Traditional stress evaluation methods, such as self-reported questionnaires and clinical interviews, often suffer from subjectivity and cannot provide continuous real-time monitoring \cite{ahmad2022survey, schmidt2018introducing}. 

Consequently, there is a growing demand for objective and automated stress recognition techniques based on physiological measurements. Among the various physiological signals used for stress assessment, electrocardiography (ECG) has emerged as one of the most reliable and widely adopted modalities because it directly reflects autonomic nervous system activity \cite{schmidt2018introducing,ahmad2023multilevel}. Changes in heart rhythm and cardiac dynamics caused by physiological and psychological stress can be effectively captured through ECG recordings. Furthermore, the increasing availability of wearable ECG monitoring devices has facilitated the deployment of continuous stress monitoring systems in both controlled and real-world environments. These advantages have motivated extensive research on ECG-based stress recognition and physiological state monitoring. 

Recent advances in deep learning have significantly enhanced the capability of automated stress classification systems by eliminating the need for handcrafted feature engineering \cite{acharya2017deep,kachuee2018ecg}. Convolutional Neural Networks (CNNs) have demonstrated remarkable effectiveness in learning discriminative local and morphological characteristics directly from raw ECG signals. Their hierarchical feature extraction process allows the network to automatically identify stress-related waveform patterns and extract informative representations without manual feature design~\cite{kiranyaz2015real}.

On the other hand, Long Short-Term Memory (LSTM) networks and their bidirectional variants (BiLSTM) are specifically designed to model temporal dependencies and sequential dynamics, making them highly suitable for analyzing the temporal behavior of ECG signals~\cite{hochreiter1997long}~\cite{schuster1997bidirectional}. Attention mechanisms further enhance recurrent networks by enabling them to focus on the most informative segments of the signal, thereby improving representation learning and classification performance. Although both CNN and BiLSTM architectures have achieved promising results individually, they learn fundamentally different yet complementary characteristics of ECG signals. The CNN primarily captures local morphological structures and waveform patterns, whereas the BiLSTM-Attention network focuses on temporal dependencies and contextual information embedded within the ECG sequence. Since stress-related cardiac responses are manifested through both morphological variations and temporal dynamics, relying on a single feature extraction paradigm may limit the representational capability of the model. Therefore, integrating complementary representations learned by these architectures has the potential to improve stress classification performance. 

Feature fusion has emerged as an effective strategy for combining heterogeneous representations and exploiting complementary information. Traditional fusion approaches commonly employ simple feature concatenation, which often fails to fully capture the interactions among different feature spaces. Current studies have demonstrated that attention-based fusion mechanisms can model relationships more effectively between heterogeneous representations through adaptive feature weighting and information exchange \cite{vaswani2017attention}. In particular, cross-attention mechanisms enable one feature representation to selectively attend to the most informative components of another representation, thereby learning richer and more discriminative embeddings. 
 
 Ongoing research on physiological signal analysis has shown that cross-attention-based fusion can effectively capture complex inter-feature relationships and enhance representation learning by exploiting complementary information from multiple feature streams~\cite{ding2025multimodal},~ \cite{khorasani2026cross}. In parallel, residual learning has proven highly effective in preserving discriminative information while facilitating deeper representation learning. By introducing shortcut connections, residual learning mitigates information degradation and enables the propagation of useful features across network layers, resulting in improved optimization and representation quality \cite{he2016deep}. The success of residual learning suggests that the preservation of original feature information during the fusion process can improve feature integration and prevent the loss of complementary characteristics. Moreover, recent research indicates that the combination of complementary fusion paradigms can provide a more effective integration mechanism than the use of a single fusion strategy. By jointly exploiting multiple forms of feature interaction, hybrid fusion frameworks can generate richer feature representations and improve classification performance.
 
  Motivated by these observations, this work investigates the fusion of complementary morphological and temporal ECG representations for stress classification. Specifically, a 1D-CNN network is used to extract morphological features from raw ECG signals, while a BiLSTM-Attention network is utilized to capture temporal dependencies and contextual information. Building upon these complementary representations, three feature fusion frameworks are explored. The first framework adopts a Residual Feature Fusion (RFF) strategy to preserve important information while promoting feature interaction. The second framework introduces Bidirectional Feature-Wise Cross-Attention Fusion (BFCAF), which enables reciprocal information exchange between CNN and BiLSTM feature representations. Finally, a novel Cross-Attention Residual Fusion (CARF) framework is proposed to combine the strengths of residual learning and bidirectional cross-attention within a unified fusion architecture. The proposed frameworks are evaluated using subject-independent experiments on benchmark ECG stress datasets. Experimental evaluation confirms that fusing complementary morphological and temporal ECG representations improves stress classification outcomes compared to models based on a single feature extraction strategy. Moreover, the proposed CARF framework achieves superior performance by simultaneously promoting effective inter-feature interaction and preserving discriminative information throughout the feature fusion process. The main contributions of this study are summarized as follows: 
  
    \begin{itemize} 
        \item A dual-branch framework is developed that combines a 1D-CNN and a BiLSTM-Attention network to learn complementary morphological and temporal representations from raw ECG signals for stress classification.
      
        \item  A residual-based fusion framework is introduced to integrate CNN and BiLSTM feature representations while preserving important information learned by the individual branches.
        
        \item A bidirectional cross-attention mechanism is developed to model interactions between CNN-derived and BiLSTM-derived feature representations, allowing each branch to selectively incorporate information from the other.
        
        \item A hybrid fusion framework is proposed that combines residual learning and bidirectional cross-attention within a unified architecture. This approach aims to preserve discriminative feature information while enabling adaptive interaction between complementary representations.
        
        \item To our knowledge, this is the first study to conduct an extensive subject-independent evaluation of residual, cross-attention, and hybrid fusion strategies for the classification of stress based on three-classes of ECG. These experiments provide a comparative evaluation of residual, cross-attention, and hybrid fusion strategies, highlighting the performance of CARF relative to individual feature extraction branches and other evaluated approaches.
    \end{itemize}

\section{Related Work}
\label{sec:relatedwork}
Early approaches to physiological stress recognition relied primarily on conventional machine learning algorithms and handcrafted features extracted from physiological signals. Hidden Markov Models (HMMs), Support Vector Machines (SVMs), Linear Discriminant Analysis (LDA), Gaussian Mixture Models (GMMs), and Random Forests have been investigated for stress recognition \cite{womack1996classification,mcduff2014remote,giakoumis2012using,kurniawan2013stress,tsiknakis2018stress}. \cite{garg2021stress} evaluated several machine learning methods for binary and three-class stress classification using the WESAD dataset \cite{schmidt2018introducing}, incorporating physiological measurements such as electrocardiogram (ECG), body temperature, respiration, electromyogram (EMG), and electrodermal activity (EDA). Although these methods established the feasibility of physiological stress recognition, their reliance on handcrafted features can limit the representation of complex physiological patterns and often requires domain-specific expertise. These limitations motivated the adoption of deep learning methods capable of automatically extracting informative representations from physiological signals.

Convolutional Neural Networks (CNNs) have been extensively investigated for extracting local morphological and temporal characteristics from physiological recordings. \cite{giannakakis2019novel} investigated multi-kernel one-dimensional CNNs for ECG-based stress recognition, demonstrating the benefits of learning signal characteristics on different receptive-field scales. Similarly, \cite{cho2019ambulatory} developed an end-to-end CNN framework for stress detection directly from raw ECG signals, while \cite{he2019real} evaluated CNN-based recognition of acute cognitive stress using short ECG segments. Beyond the ECG,  \cite{huynh2021stressnas} explored optimized deep learning architectures for stress classification using wrist-worn physiological measurements. Multimodal approaches have also been investigated to incorporate complementary physiological and behavioral information. \cite{seo2022deep} examined the integration of ECG, respiratory and facial information at the feature-level and at the decision-level for the detection of work-related stress. Related investigations involving virtual reality environments further demonstrated the applicability of physiological signal analysis to stress recognition under controlled experimental conditions \cite{vaitheeshwari2022stress,ahmad2023multilevel,ahmad2025stress}. These studies establish convolutional feature learning as a promising approach to physiological stress recognition while motivating the incorporation of temporal dependencies and complementary signal representations.

To capture sequential dependencies beyond localized signal patterns, researchers have integrated convolutional feature extraction with recurrent architectures, particularly Long Short-Term Memory (LSTM) and Bidirectional LSTM (BiLSTM) networks. Recent investigations have incorporated attention mechanisms to emphasize temporal information relevant to the task. \cite{bipro2026attention} introduced an attention-driven architecture for the recognition of stress from single-channel ECG signals, combining convolutional feature extraction with recurrent sequence modeling. \cite{rahul2025stress} investigated CNN, LSTM, and attention-based learning for stress detection using physiological signals from WESAD. Similarly, \cite{perumalsamy2026ai} investigated multiple learning models for stress detection using physiological dynamics, contributing to the growing evidence on the role of temporal information in stress recognition. In related cardiac classification applications, \cite{najia2025enhanced} combined CNN, BiLSTM, and attention mechanisms for the classification of ECG segments, while \cite{sharma2026attention} investigated an attention-enhanced CNN-BiLSTM framework with confidence-gated inference for the detection of arrhythmias. The applicability of convolutional-recurrent learning also extends to other physiological signals, as demonstrated by \cite{kumari2025advanced} in the recognition of epileptic seizures based on EEG. These investigations establish the relevance of combining local feature extraction with sequential modeling, although their architectural integration does not necessarily establish the individual contributions or complementarity of the resulting representations.

Attention-based feature fusion has emerged as an important strategy for integrating information from multiple signal channels, feature extraction branches, and physiological modalities. \cite{sharan2025cardiac} proposed a multi-branch CNN with attention-based feature fusion for cardiac abnormality detection using 12-lead ECG signals. \cite{abidi2026multi} developed a dual-stream CNN-LSTM network that incorporates mid-level fusion and temporal attention for multimodal stress classification. Similarly, \cite{ju2026differential} introduced a differential channel-assisted CNN-LSTM network with attention fusion for the automated diagnosis of myocardial infarction. Extending this direction, \cite{dong2026multimodal} investigated a CNN-LSTM network based on cross-attention for the classification of arrhythmias, illustrating the integration of complementary ECG representations through attention-guided interactions. \cite{wu2026novel} further explored adaptive CNN-LSTM fusion for ECG diagnosis, emphasizing the relevance of integrating convolutional and recurrent features. These studies demonstrate the growing diversity of feature integration mechanisms, ranging from attention-weighted feature aggregation to adaptive and cross-attention-based fusion. However, attention-based weighting within individual representations and attention-mediated fusion between independently learned representations represent distinct architectural operations that warrant separate evaluation.

Current Studies have also explored complementary feature learning through multibranch processing, residual connections, and bidirectional attention. \cite{peng2026ecg} introduced a beat-level complementary feature fusion approach for ECG-based emotion recognition, combining morphological information and beat-level variations through separate feature learning pathways. In particular, this work provides a closely related precedent for investigating complementary ECG representations under subject-independent evaluation. \cite{kotegar2026cares} developed CaReS-BiNet, which incorporates multiscale convolutional processing and bidirectional temporal modeling for the classification of ECG arrhythmias. \cite{hassoon2026odual} investigated dual-branch bidirectional attention fusion of temporal and hierarchical representations, providing a particularly relevant architectural comparison to integrate distinct ECG feature streams. Beyond the ECG-only classification, \cite{qu2026maf} proposed MAF-Net, a multimodal cross-attention-based fusion network for the classification of cardiovascular diseases. Collectively, these studies establish that complementary feature learning, residual processing, and integration based on bidirectional attention have already been explored in ECG analysis. Nevertheless, differences in feature representations, fusion objectives, and classification settings motivate systematic comparisons to determine the relative effectiveness of these mechanisms for the recognition of single-channel ECG stress.

Integration of Attention-guided features has also been investigated in broader physiological sensing applications. \cite{11463684} proposed an explainable attention-based LSTM fusion network for the prediction of cognitive load using multimodal physiological signals. \cite{hyung2025deepattnet} investigated cross-attention mechanisms for subject-independent mental stress detection using bilateral ear-EEG signals, while \cite{swapna2026adaptive} examined adaptive multi-domain attention for single-channel EEG classification. \cite{junior2026multimodal} conducted an ablation study of EEG-ECG fusion strategies for emotion recognition, examining the contribution of multimodal integration to classification performance. These investigations provide further evidence of the applicability of attention mechanisms to learning of physiological representation and feature integration. However, differences in sensing modalities, prediction targets, and experimental protocols limit direct comparisons between these applications. In particular, improvements obtained through multichannel or multimodal fusion do not necessarily establish equivalent benefits when integrating independently learned morphological and temporal representations from a single-channel ECG signal.

Beyond architectural developments, recent research has emphasized the quality, robustness, and generalization of representation in subjects and acquisition conditions. \cite{sarkar2020self} investigated self-supervised learning for ECG-based emotion recognition, while contrastive representation learning has been explored to improve the transferability of physiological representations \cite{yang2025contrastive}. Related studies have examined personalized domain adaptation \cite{xu2026variational}, probability calibration \cite{akkaya2026calibration}, and integration of domain knowledge with deep feature learning \cite{birkenmaier2026hybrid}. Investigations involving cross-population and cross-activity evaluation have further highlighted the challenges associated with physiological variability and changes in signal distributions \cite{sahu2025anxiety,kwon2026wellness}. Recent reviews similarly identify dataset heterogeneity, limited external validation, and inconsistent evaluation protocols as persistent challenges in the recognition of wearable stress \cite{sosa2026beyond,pareja2026machine,li2025multimodal,khandelwal2025machine,sharma2026machine}. These findings underscore the importance of subject-independent evaluation and consistent experimental protocols when assessing physiological classification models. Furthermore, \cite{kiymik2026dual} investigated attention-based fusion against an alternative fusion strategy in low-label ECG-PCG classification, providing a relevant methodological precedent for evaluating whether more complex fusion mechanisms offer measurable benefits over simpler alternatives.

Despite advances in convolutional-recurrent modeling and attention-based feature fusion, the comparative effectiveness of different mechanisms to integrate independently learned morphological and temporal representations into single-channel ECG stress classification remains an important research question. Existing studies have investigated complementary ECG feature fusion, residual learning, adaptive integration, and bidirectional attention in related physiological classification tasks \cite{peng2026ecg,dong2026multimodal,wu2026novel,hassoon2026odual,qu2026maf}. However, differences in representation design, fusion mechanisms, datasets, and evaluation protocols make it difficult to isolate the contributions of individual fusion strategies. Although subject-independent stress recognition and complementary feature fusion have previously been investigated, systematic comparisons of residual and bidirectional attention-based fusion mechanisms under consistent multi-class, single-channel ECG evaluation remain valuable. Motivated by these considerations, this study investigates three feature fusion frameworks, RFF, BFCAF, and CARF, incorporating residual learning and bidirectional cross-attention to integrate complementary convolutional and recurrent ECG representations. The proposed frameworks are systematically evaluated using leave-one-subject-out cross-validation (LOSOCV) across two benchmark ECG stress datasets to examine their relative effectiveness in feature discrimination, multi-class classification, and generalization to unseen subjects.

\section{Proposed Feature Fusion Methods} \label{ sec : Proposed Methods}
	
In this section, we introduce proposed deep learning based feature fusion methods used to integrate the complementary information captured by the 1D-CNN and BiLSTM-Attention networks. 1D-CNN extracts morphological feature representation from the ECG signal, while the BiLSTM-Attention network extracts temporal feature representation. These feature vectors are subsequently fused to produce a unified representation for stress classification.

	\begin{figure}[h]
    \centering
    	\includegraphics[width=\linewidth]{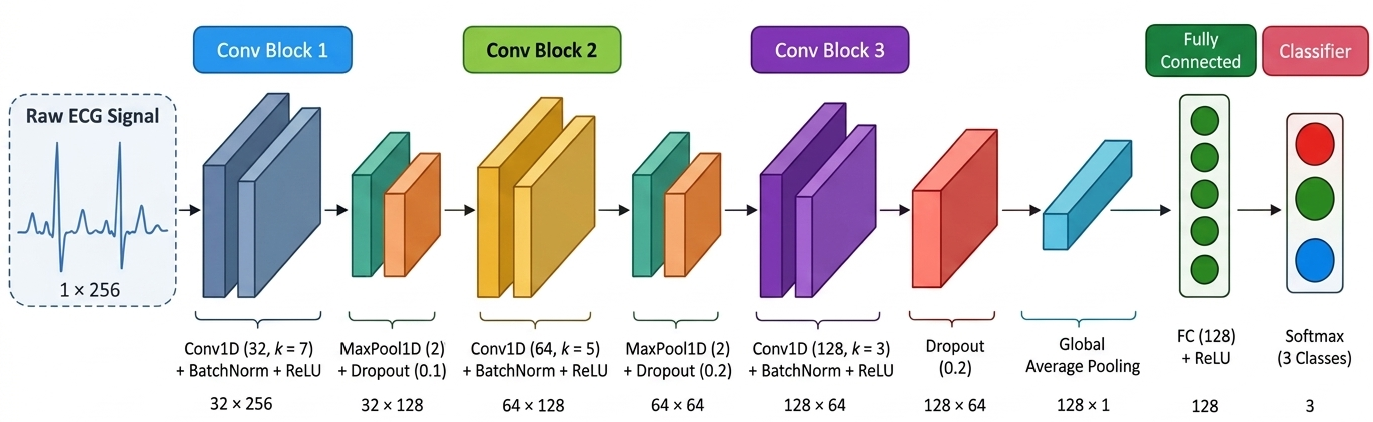}
    	\caption{Architecture of the proposed 1D-CNN model for ECG feature extraction and classification.}
    	\label{fig:1DCNN Architecture}
    \end{figure}
    
    \begin{figure}[h]
    \centering
    	\includegraphics[width=\linewidth]{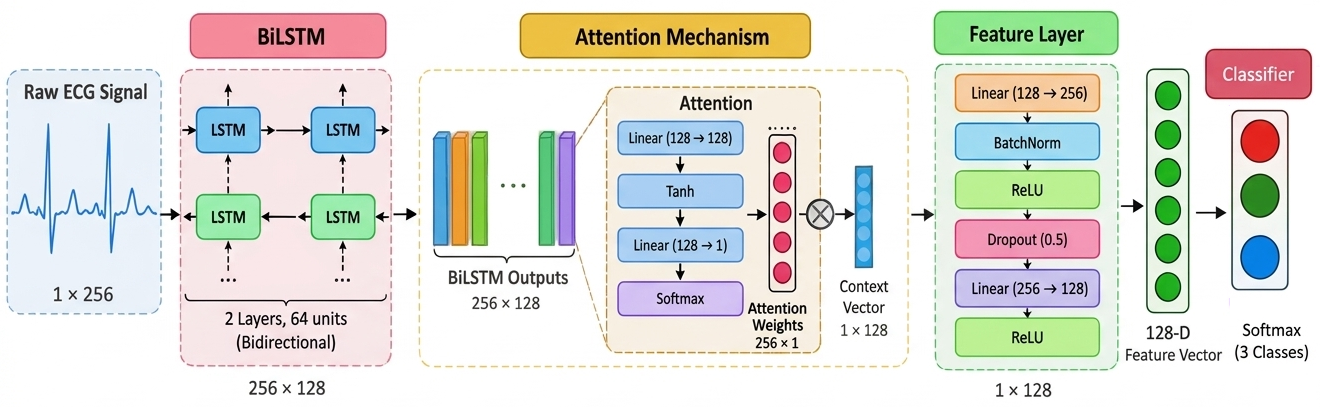}
    	\caption{Architecture of the proposed BiLSTM-Attention Architecture for ECG feature extraction and classification.}
    	\label{fig:LSTM Architecture}
    \end{figure}

    \begin{figure}[h]
	\centering
    	\includegraphics[width=\linewidth]{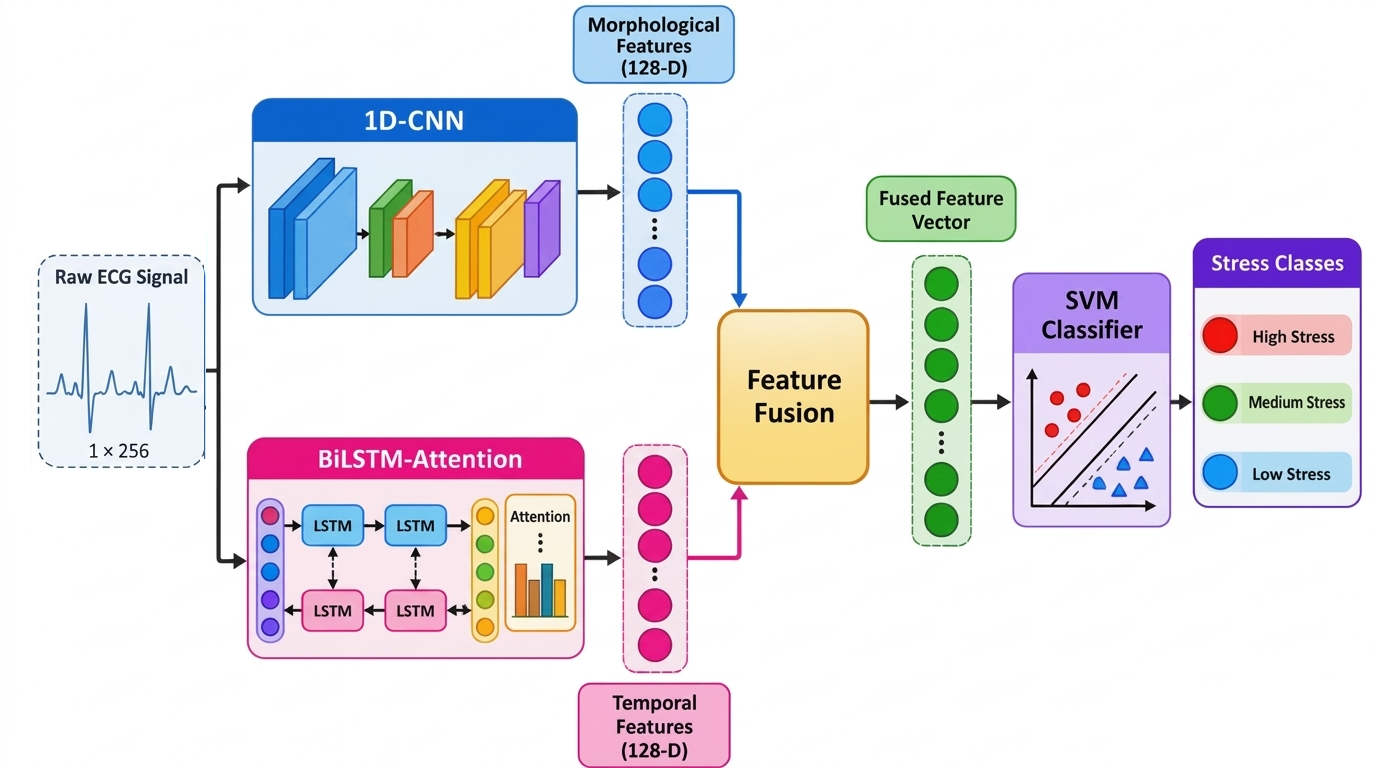}
    	\caption{Overview of the proposed fusion-based stress classification framework. Morphological features extracted from the 1D-CNN and temporal features extracted from the BiLSTM-Attention network are combined through a feature fusion module to construct a unified feature representation for SVM-based stress classification.}
    	\label{fig: Overview of proposed feature fusion}
    \end{figure}

The overall architecture of the proposed fusion-based stress classification framework is shown in Fig~\ref{fig: Overview of proposed feature fusion}. The raw ECG signal is processed in parallel by a 1D-CNN and a BiLSTM-Attention network to capture complementary morphological and temporal characteristics, respectively. These complementary feature vectors are then integrated through a feature fusion module to generate an integrated feature representation. The fused feature vector is subsequently provided to an SVM classifier for stress level prediction.

The details of each fusion method are provided below.

\subsection{Residual Feature Fusion}

 Combining complementary features extracted from multiple network branches can significantly improve physiological signal classification performance. In ~\cite{han2024online} a multibranch multiscale co-learning framework for ECG classification is proposed. This framework highlights the effectiveness of integrating representations learned from different feature extraction pathways. Similarly, in ~\cite{kotegar2026cares}, multi-scale deep feature aggregation is proposed to capture complementary arrhythmia characteristics. Furthermore, for multimodal physiological signal analysis, a graph residual fusion mechanism is designed inc \cite{zhao2026grf} to enhance feature interaction and information preservation across modalities. Also, the work in ~\cite{lee2023seriessleepnet} demonstrated the importance of combining diverse temporal representations in biosignal learning tasks.
 
  Motivated by these advances, in this work, we propose a residual feature fusion strategy to integrate complementary CNN and BiLSTM representations extracted from ECG signals.
  
  Let $x_{CNN} \in \mathbb{R}^{d}$ and $x_{BiLSTM} \in \mathbb{R}^{d}$ denote the feature vectors extracted by the CNN and BiLSTM branches, respectively. The two deep networks capture distinct yet complementary aspects of ECG signals. The 1D-CNN extracts local morphological features, while the BiLSTM-Attention network learns temporal dependencies, thereby providing complementary representations for feature fusion as shwon in Fig~\ref{fig: Overview of proposed feature fusion}. To exploit the complementary information provided by both representations, the two feature vectors are first concatenated as 
  \begin{equation} x_F = [x_{CNN}; x_{BiLSTM}] \label{eq:concat} \end{equation} 
  where $[\cdot;\cdot]$ denotes the concatenation operation. The concatenated representation is then processed through a residual transformation network consisting of two fully connected layers with ReLU activation and dropout regularization 
  \begin{equation} r = W_2 \Big( \mathrm{Dropout} \big( \mathrm{ReLU}(W_1x_F + b_1) \big) \Big) + b_2 \label{eq:residual1} \end{equation} 
  where $W_1 \in \mathbb{R}^{h \times 2d}$ and $W_2 \in \mathbb{R}^{d \times h}$ are trainable weight matrices, $b_1$ and $b_2$ are bias vectors, and $h$ denotes the hidden-layer dimension. Unlike conventional concatenation-based fusion methods, the proposed approach learns an additional residual representation that captures higher-order interactions between CNN and LSTM features. The final fused representation is obtained through residual addition 
  \begin{equation} f = x_{CNN} + x_{BiLSTM} + r \label{eq:fusion} \end{equation} 
  where $f$ denotes the final fused feature vector. Substituting Equation~(\ref{eq:residual1}) into Equation~(\ref{eq:fusion}) yields the complete residual fusion formulation
   \begin{equation} f = x_{CNN} + x_{BiLSTM} + W_2 \Big( \mathrm{Dropout} \big( \mathrm{ReLU} ( W_1[x_{CNN};x_{BiLSTM}] +b_1 ) \big) \Big) + b_2 \label{eq:complete_fusion} \end{equation} 
   This residual fusion mechanism preserves the original information contained in both CNN and LSTM features while simultaneously learning complementary discriminative patterns from their joint representation. The resulting fused feature vector is subsequently used for classification. In the proposed framework, the residual fusion network is first trained using a fully connected classifier and cross-entropy loss. After training, the fused representation $f$ is extracted and employed as the input to a Support Vector Machine (SVM) classifier with a radial basis function (RBF) kernel for final ECG classification. This two-stage strategy enables the network to learn discriminative fused embeddings while leveraging the strong generalization capability of SVMs.

\subsection{Bidirectional Feature-Wise Cross-Attention Fusion} Recent advances in ECG signal classification have demonstrated the effectiveness of attention-based feature fusion techniques for combining complementary feature representations extracted from multiple learning branches.
In ~\cite{dong2026multimodal}, a cross-attention-based CNN--LSTM architecture is employed to facilitate interaction between heterogeneous feature representations for arrhythmia classification. Similarly, ~\cite{qu2026maf} introduced a bidirectional cross-attention fusion mechanism for cardiovascular disease classification, demonstrating the benefits of reciprocal feature interaction. The studies in ~\cite{deng2025efficientecg} further showed that cross-attention-based feature fusion can enhance ECG classification performance by effectively integrating multiple information sources. In addition, the research in ~\cite{lee2023seriessleepnet} highlighted the value of combining CNN and recurrent features through learnable fusion strategies in physiological signal analysis.

 Inspired by these studies, in this work, we propose our second deep learning based feature fusion framework called Bidirectional Feature-Wise Cross-Attention Fusion (BFCAF) to integrate complementary CNN and LSTM feature representations extracted from ECG signals.
 
  Let $x_{CNN}\in\mathbb{R}^{d}$ and $x_{BiLSTM}\in\mathbb{R}^{d}$ denote the feature vectors generated by the CNN and LSTM branches, respectively, where $d$ represents the feature dimensionality. The CNN branch primarily captures local morphological characteristics of ECG waveforms, whereas the LSTM branch models temporal and sequential dependencies. The objective of the proposed BFCAF module is to enable bidirectional interaction between these two complementary feature streams before the final classification stage.
  
   It should be noted that the proposed BFCAF module differs fundamentally from the attention mechanism employed within the BiLSTM feature extraction network. The attention layer in the BiLSTM branch operates on the ECG sequence to identify temporally important samples and generate a more informative temporal representation as explained in Section ~\ref{sec : BiLSTM}. In contrast, BFCAF operates after feature extraction and performs feature-level interaction between the high-level CNN and LSTM representations. Therefore, the attention mechanism within the BiLSTM enhances temporal feature learning, whereas BFCAF enhances inter-stream feature fusion. Unlike conventional Transformer cross-attention, which computes attention using matrix multiplication between query and key representations, the proposed method employs feature-wise attention based on element-wise query--key interaction. This design maintains bidirectional information exchange while reducing computational complexity and preserving interpretability. 

\subsubsection{CNN-to-LSTM Feature-Wise Attention} In the first attention direction, the CNN representation acts as the query, whereas the LSTM representation provides the key and value vectors: \begin{equation} Q_{CNN}=W_Q^{C}x_{CNN} \label{eq:qcnn} \end{equation} \begin{equation} K_{BiLSTM}=W_K^{L}x_{BiLSTM} \label{eq:klstm} \end{equation} \begin{equation} V_{BiLSTM}=W_V^{L}x_{BiLSTM} \label{eq:vlstm} \end{equation} where $W_Q^{C}$, $W_K^{L}$, and $W_V^{L}$ denote learnable projection matrices. The feature-wise attention scores are computed as \begin{equation} S_{CNN\rightarrow BiLSTM} = \frac{ Q_{CNN}\odot K_{BiLSTM} } {\sqrt{d}} \label{eq:score_cnn} \end{equation} where $\odot$ denotes element-wise multiplication. The normalized attention weights are obtained through a softmax operation: \begin{equation} A_{CNN\rightarrow BiLSTM} = \mathrm{Softmax} \left( S_{CNN\rightarrow BiLSTM} \right) \label{eq:attn_cnn} \end{equation} The resulting cross-attended representation is \begin{equation} Z_{CNN\rightarrow BiLSTM} = A_{CNN\rightarrow BiLSTM} \odot V_{BiLSTM} \label{eq:zcnn} \end{equation} This operation enables CNN-derived information to selectively emphasize informative components of the LSTM representation. 

\subsubsection{LSTM-to-CNN Feature-Wise Attention} To establish reciprocal information exchange, a second attention branch is implemented in the opposite direction. In this branch, the LSTM representation generates the query, whereas the CNN representation provides the key and value vectors: \begin{equation} Q_{LSTM}=W_Q^{L}x_{BiLSTM} \label{eq:qlstm} \end{equation} \begin{equation} K_{CNN}=W_K^{C}x_{CNN} \label{eq:kcnn} \end{equation} \begin{equation} V_{CNN}=W_V^{C}x_{CNN} \label{eq:vcnn} \end{equation} The attention weights are computed as \begin{equation} A_{BiLSTM\rightarrow CNN} = \mathrm{Softmax} \left( \frac{ Q_{LSTM}\odot K_{CNN} } {\sqrt{d}} \right) \label{eq:attn_lstm} \end{equation} The cross-attended CNN representation is then obtained by \begin{equation} Z_{BiLSTM\rightarrow CNN} = A_{BiLSTM\rightarrow CNN} \odot V_{CNN} \label{eq:zlstm} \end{equation} Consequently, the LSTM representation selectively highlights informative components of the CNN representation, providing complementary interaction to the CNN-to-LSTM branch.

 \subsubsection{Cross-Attention Feature Fusion} The two cross-attended representations are concatenated to form a joint feature vector: \begin{equation} F= \left[ Z_{CNN\rightarrow BiLSTM}; Z_{BiLSTM\rightarrow CNN} \right] \label{eq:concat_fusion} \end{equation} The concatenated representation is subsequently projected through a fully connected transformation followed by ReLU activation and dropout regularization: \begin{equation} f = \mathrm{Dropout} \Big( \mathrm{ReLU} ( W_fF+b_f ) \Big) \label{eq:final_fusion} \end{equation} where $W_f$ and $b_f$ denote the trainable parameters of the fusion layer. The resulting vector $f$ represents the final fused feature representation and has a dimensionality of 128 in the implemented architecture.
 
  \subsubsection{Classification} For network training, the fused representation is forwarded through a fully connected classifier and optimized using the cross-entropy loss function. \begin{equation} \hat{y} = W_cf+b_c \label{eq:classifier} \end{equation} where $\hat{y}$ denotes the classifier logits. After training, the learned fused feature representation $f$ is extracted and employed as the input to a Support Vector Machine (SVM) classifier with a radial basis function (RBF) kernel for final three-class ECG classification.
  
   \subsubsection{Methodological Interpretation} 
   
   The proposed BFCAF mechanism enables reciprocal interaction between CNN and LSTM feature streams before the final fusion stage. In the CNN-to-LSTM branch, CNN-derived information selectively emphasizes informative components of the LSTM representation, whereas the LSTM-to-CNN branch performs the complementary operation. Unlike attention mechanisms operating within the feature extraction networks, BFCAF acts on the high-level representations generated by both branches and therefore focuses on inter-stream feature fusion rather than temporal feature weighting. The complete BFCAF operation can be summarized as \begin{equation} f = \phi \Big( [ (A_{CNN\rightarrow BiLSTM}\odot V_{BiLSTM}); (A_{BiLSTM\rightarrow CNN}\odot V_{CNN}) ] \Big) \label{eq:overall_bfcaf} \end{equation} where $\phi(\cdot)$ denotes the nonlinear fusion transformation defined in Eq.~(\ref{eq:final_fusion}).

\subsection{Cross-Attention Residual Fusion}
The results of residual fusion and BFCAF are shown in Tables~\ref{tab : Residual Fusion on RML}and ~\ref{tab : Cross Attention Fusion on RML}. The success of two proposed feature fusion methods encouraged us to investigates the integration of cross-attention and residual learning within a unified fusion framework. Furthermore, the experiments performed on physiological signals in ~\cite{kumar2026trust} proved that combination of two fusion frameworks significantly improve the performance of stress classification using physiological signals. Thus, we introduce our novel deep learning based feature fusion framework called  Cross-Attention Residual Fusion (CARF). This framework simultaneously exploit bidirectional feature interaction and information-preserving residual learning.
 
Let $x_{CNN}\in\mathbb{R}^{d}$ and $x_{BiLSTM}\in\mathbb{R}^{d}$ denote the feature vectors extracted from the CNN and LSTM branches, respectively. The CNN branch primarily captures local morphological characteristics of ECG waveforms, whereas the LSTM branch learns temporal and sequential dependencies. It should be noted that the proposed CARF module differs from any attention mechanism employed within the LSTM feature extractor. Attention within the LSTM branch operates on ECG sequences to identify important temporal regions, whereas the proposed CARF module operates on the high-level feature representations generated by both branches and therefore focuses on inter-representation fusion. Unlike conventional scaled dot-product cross-attention, the proposed method employs feature-wise query--key interactions through element-wise multiplication. Accordingly, the CNN-to-LSTM attention branch is computed as \begin{equation} Z_{CNN\rightarrow BiLSTM} = \mathrm{Softmax} \left( \frac{ (W_Q^{C}x_{CNN}) \odot (W_K^{L}x_{BiLSTM}) } {\sqrt{d}} \right) \odot (W_V^{L}x_{BiLSTM}) \label{eq:cnn_lstm} \end{equation} where $\odot$ denotes element-wise multiplication. Similarly, the reciprocal LSTM-to-CNN interaction is defined as \begin{equation} Z_{BiLSTM\rightarrow CNN} = \mathrm{Softmax} \left( \frac{ (W_Q^{L}x_{BiLSTM}) \odot (W_K^{C}x_{CNN}) } {\sqrt{d}} \right) \odot (W_V^{C}x_{CNN}) \label{eq:lstm_cnn} \end{equation} where $W_Q$, $W_K$, and $W_V$ denote trainable projection matrices. The resulting cross-attended representations are concatenated and projected through a nonlinear transformation. \begin{equation} F_{P} = \mathrm{ReLU} \left( W_{P} \left[ Z_{CNN\rightarrow BiLSTM}; Z_{BiLSTM\rightarrow CNN} \right] +b_{P} \right) \label{eq:projection} \end{equation}

where $F_P$ represents the learned attention-enhanced interaction features. The projection matrix $W_P$ is designed such that $F_P \in \mathbb{R}^{d}$, ensuring dimensional compatibility with $x_{CNN}$ and $x_{BiLSTM}$ for the residual addition in Eq.~(\ref{eq:residual}). To preserve information independently learned by the CNN and LSTM branches, residual fusion is employed.

Unlike RFF, where the residual component is obtained through a nonlinear transformation of the concatenated CNN and BiLSTM features, CARF utilizes the attention-enhanced interaction features $F_P$ as the residual term. This enables the model to preserve the original feature representations while incorporating complementary information learned through bidirectional cross-attention.

\begin{equation} F_{R} = x_{CNN} + x_{BiLSTM} + F_{P} \label{eq:residual} \end{equation} This residual formulation enables the model to simultaneously utilize the original CNN representation, the original LSTM representation, and the learned cross-attention interaction information. The fused representation is subsequently transformed into a compact discriminative feature space through a fully connected layer, followed by ReLU activation and dropout regularization. \begin{equation} f = \mathrm{Dropout} \Big( \mathrm{ReLU} ( W_{F}F_{R}+b_{F} ) \Big) \label{eq:final_feature} \end{equation} where $f\in\mathbb{R}^{d}$ denotes the final fused feature representation. During network optimization, the learned features are supplied to a temporary neural classifier and trained using the cross-entropy loss. After training, the d-dimensional representation $f$ is extracted and provided to a Support Vector Machine (SVM) classifier with a radial basis function (RBF) kernel for final ECG classification. The proposed CARF framework provides three primary advantages. First, it combines complementary morphological and temporal representations learned by CNN and LSTM networks. Second, it performs bidirectional cross-attention interaction, allowing each feature stream to selectively emphasize informative components of the other. Third, the residual fusion formulation preserves the original feature information while simultaneously incorporating learned interaction features, thereby reducing the risk of information loss during fusion.

\section{Experiments and Results} \label{ sec : Experimental results}
We conducted experiments on the RML dataset~\cite{ahmad2023multilevel} and the WESAD dataset~\cite{schmidt2018introducing} for three-class stress classification. For both datasets, model performance was evaluated using the leave-one-subject-out cross-validation (LOSOCV) protocol. In LOSOCV, data from one subject are reserved for testing, while data from all remaining subjects are used for training. This process is repeated until every subject has served as the test subject exactly once, and the final performance is reported as the average across all folds~\cite{ahmad2022survey}. Unlike conventional cross-validation methods, LOSOCV provides a more realistic assessment of subject-independent performance by evaluating the model on entirely unseen individuals. Consequently, this evaluation strategy is particularly suitable for real-world stress monitoring applications, where the system is expected to generalize effectively to new users without requiring subject-specific retraining.

\subsection{RML Dataset}

For the RML dataset, nine subjects participated in a controlled experimental protocol designed to induce varying levels of stress while physiological signals were continuously recorded. The collected signals included electrocardiogram (ECG), galvanic skin response (GSR), and respiration data. ECG recordings were acquired using a chest-mounted sensor that measured the electrical activity of the heart. All physiological signals were sampled at a frequency of 256 Hz. In the present study, only the ECG modality was utilized for stress assessment. Focusing exclusively on ECG provides a practical and cost-effective solution, as ECG signals can be readily acquired using commercially available wearable devices, including modern smartwatches and fitness trackers. The RML dataset comprises three stress categories: Low Stress, Medium Stress, and High Stress, which were used as the target classes for the stress classification experiments.

\subsubsection{Baseline Experiments}

To assess the effectiveness of the proposed deep learning based fusion frameworks explained in Section~\ref{ sec : Proposed Methods}, baseline experiments were first conducted on the RML dataset. The first baseline model employed a 1D-CNN architecture trained directly on raw ECG signals to perform three-class stress classification.

\paragraph{1D-CNN Model Architecture and Hyperparameter Settings}

The proposed 1D-CNN feature extractor consists of three convolutional blocks with 32, 64, and 128 filters and kernel sizes of 7, 5, and 3, respectively. Each convolutional layer is followed by batch normalization and a ReLU activation function to improve training stability and introduce non-linearity. Max-pooling layers are employed after the first two convolutional blocks to reduce the temporal dimensionality of the feature maps, while dropout layers are incorporated to mitigate overfitting. The extracted feature maps are subsequently aggregated using global average pooling and passed through a fully connected layer with 128 neurons to obtain a compact feature representation. Finally, a softmax classification layer is used for stress classification.

The training hyperparameters employed for the 1D-CNN model are summarized in Table~\ref{tab:1dcnn_hyperparameters}. The optimal hyperparameter configuration was determined through grid search with cross-validation, and the parameter combination yielding the best validation performance was selected for model training. The results with 1D CNN are shown in Table~\ref{tab : 1D RML Dataset}.

\begin{table}[h] 
    \centering
    \begin{tabular}{c c} 
        \hline
        \textbf{Hyperparameter} & \textbf{Value} \\
        \hline  Optimizer & Adam \\
        \hline  Learning Rate & 0.001 \\
        \hline  Loss Function & Cross-Entropy Loss \\
        \hline  LR Scheduler & ReduceLROnPlateau \\
        \hline  LR Decay Factor & 0.5 \\
        \hline  LR Scheduler Patience & 10 \\
        \hline  Minimum Learning Rate & $1\times10^{-6}$ \\
        \hline  Early Stopping Patience & 10 \\
        \hline  Early Stopping Delta & 0.001 \\
        \hline  Batch Size & 32 \\
        \hline  
    \end{tabular}
    \caption{Hyperparameter settings used for training the 1D-CNN model.} 
    \label{tab:1dcnn_hyperparameters} 
\end{table}

\begin{table}[h]
    \centering
    \begin{tabular}{c c c c c}	
        \hline
        \textbf{Testing Sub} & \textbf{\makecell{Accuracy}} & \textbf{\makecell{Precision}} & \textbf{\makecell{Recall}} & \textbf{$F_1$ Score} \\
        \hline 
        2 & 72.8 & 81.3 & 71.6 & 70.8 \\
        \hline 
        3 & 79.3 & 87 & 79.1 & 79.8  \\
        \hline 
        4 & 79.3 & 86.4.4 & 79.1 & 79.5  \\
        \hline 
        5 & 77.5 & 83.7 & 77.3 & 77.7  \\
        \hline 
        11 & 47.6 & 29.9 & 47.2 & 36.15  \\
        \hline 
        12 & 74.3 & 83.6 & 73 & 73.7  \\
        \hline 
        13& 78.1 & 86.9 & 77.7 & 78.4  \\
        \hline 
        14& 77.1 & 86.3 & 76.1 & 77.1  \\
        \hline 
        16 & 76.2 & 84.6 & 75.6 & 76 \\
        \hline 
        \textbf{Average} & \textbf{73.6} & \textbf{78.9} & \textbf{73} & \textbf{72.1} \\
        \hline 
    \end{tabular}
	\caption{Performance metrics with 1DCNN on RML dataset}
	\label{tab : 1D RML Dataset}
\end{table} 

\paragraph{BiLSTM-Attention Architecture and Training Configuration}\label{sec : BiLSTM}

The proposed BiLSTM-Attention network was designed to capture the temporal dynamics of ECG signals and generate discriminative deep features. The model consists of a two-layer Bidirectional Long Short-Term Memory (BiLSTM) network with 64 hidden units in each direction. By processing the input sequence in both forward and backward directions, the BiLSTM captures contextual information from past and future time steps simultaneously. To emphasize the most informative segments of the ECG signal, an attention mechanism is applied to the BiLSTM outputs. The attention module comprises a fully connected layer followed by a hyperbolic tangent activation function and a scoring layer that assigns importance weights to each temporal feature representation. The weighted hidden states are then aggregated to form a context vector that summarizes the relevant information contained in the ECG sequence.

The resulting context vector is passed through a feature extraction module consisting of two fully connected layers. The first layer transforms the 128-dimensional attention-based representation into 256 feature units, followed by batch normalization, ReLU activation, and dropout regularization to improve generalization performance. A second fully connected layer reduces the feature dimensionality to 128, producing a compact deep feature representation. Finally, a softmax classification layer is employed for stress classification.

The architecture of the proposed BiLSTM-Attention model is illustrated in Fig~\ref{fig:LSTM Architecture}, while the corresponding training hyperparameters are reported in Table~\ref{tab:bilstm_hyperparameters}. The results with BiLstm are shown in Table~\ref{tab : LSTM on RML Dataset}
\begin{table}[h] 
\centering
    \begin{tabular}{c c} 
        \hline
        \textbf{Hyperparameter} & \textbf{Value} \\
        \hline 
        Loss Function & CrossEntropyLoss \\
        \hline
        Optimizer & Adam \\
        \hline
        Learning Rate & 0.0005 \\
        \hline 
        Weight Decay & 0.0005 \\
        \hline 
        Learning Rate Scheduler & ReduceLROnPlateau \\
        \hline  
        Scheduler Mode & min \\
        \hline  
        LR Reduction Factor & 0.5 \\
        \hline  
        Scheduler Patience & 10 \\
        \hline  
        Scheduler Threshold & 0.001 \\
        \hline 
        Minimum Learning Rate & $1\times10^{-6}$ \\
        \hline 
        Early Stopping Patience & 10 \\
        \hline 
        Early Stopping Min Delta & 0.001 \\
        \hline  
        Gradient Clipping & Enabled \\
        \hline  
        Gradient Clipping Max Norm & 1.0 \\
        \hline  
    \end{tabular}
\caption{Hyperparameter settings used for training the BiLSTM-Attention model.} \label{tab:bilstm_hyperparameters} 
\end{table}

\begin{table}[h]
    \centering
    \begin{tabular}{c c c c c}	
        \hline
        \textbf{Testing Sub} & \textbf{Accuracy (\%)} & \textbf{Precision (\%)} & \textbf{Recall (\%)} & \textbf{$F_1$ Score (\%)} \\
        \hline 
        2 & 71.9 & 79.3 & 70.6 & 69.7 \\
        \hline 
        3 & 75.3 & 77 & 75.2 & 75.4  \\
        \hline 
        4 & 72.7 & 78.6 & 72.6 & 71.3  \\
        \hline 
        5 & 67.9 & 72.8 & 67.4 & 64.7  \\
        \hline 
        11 & 44.1 & 52.1 & 41.3 & 34.3  \\
        \hline 
        12 & 72.5 & 80.6 & 71.2 & 71.5  \\
        \hline 
        13& 71.8 & 80.2 & 71.2 & 70  \\
        \hline 
        14& 69.7 & 75.6 & 67.9 & 66  \\
        \hline 
        16 & 74.7 & 83 & 73.9 & 73.7 \\
        \hline 
        \textbf{Average} & \textbf{68.9} & \textbf{75.5} & \textbf{67.9} & \textbf{66.3} \\
        \hline 
    \end{tabular}
	\caption{Performance metrics with BiLSTM on RML dataset}
	\label{tab : LSTM on RML Dataset}
\end{table} 

In addition to evaluating the individual models, several baseline experiments were performed using different feature fusion strategies. These experiments aimed to assess the extent to which combining the complementary feature representations learned by the 1D-CNN and BiLSTM-Attention models could improve stress classification performance.

\paragraph{Feature Level Concatenation Fusion}

The complementary features learned by the 1D-CNN and BiLSTM-Attention models were integrated using a feature-level concatenation strategy. The 1D-CNN extracts morphological characteristics of the ECG signal, whereas the BiLSTM-Attention network captures temporal dependencies and contextual information. Let $\mathbf{f}_{CNN} \in \mathbb{R}^{128}$ and $\mathbf{f}_{LSTM} \in \mathbb{R}^{128}$ denote the feature vectors generated by the two networks, respectively. These feature representations were concatenated along the feature dimension to form a unified 256-dimensional feature vector: \begin{equation} \mathbf{f}_{fusion} = [\mathbf{f}_{CNN};\mathbf{f}_{LSTM}] \in \mathbb{R}^{256} \end{equation} where $[\ ;\ ]$ denotes the concatenation operation. The resulting fused feature vector combines both morphological and temporal information, enabling a more comprehensive representation of the ECG signal.  Finally, an SVM classifier was trained using the fused feature vectors for stress classification. The results with concatenation fusion are shown in Table~\ref{tab : Concatenation Fusion on RML}

\begin{table}[h]
    \centering
    \begin{tabular}{c c c c c}	
        \hline
        \textbf{Testing Sub} & \textbf{Accuracy (\%)} & \textbf{Precision (\%)} & \textbf{Recall (\%)} & \textbf{$F_1$ Score (\%)} \\
        \hline 
        2 & 79.7 & 87.6 & 79.4 & 80.2 \\
        \hline 
        3 & 84.4 & 86.1 & 84.1 & 84.4  \\
        \hline 
        4 & 85.9 & 90.1 & 85.7 & 86  \\
        \hline 
        5 & 85.9 & 87.5 & 85.9 & 85.6  \\
        \hline 
        11 & 73.1 & 82.3 & 71.7 & 73 \\
        \hline 
        12 & 75.9 & 82.8 & 74.3 & 74.3  \\
        \hline 
        13& 79.7 & 84.5 & 79.7 & 80.2 \\
        \hline 
        14& 79.7 & 87.6 & 79.3 & 80.2  \\
        \hline 
        16 & 79.8 & 87.7 & 79.5 & 80.3 \\
        \hline 
        \textbf{Average} & \textbf{80.5} & \textbf{86.2} & \textbf{80} & \textbf{80.5} \\
        \hline 
				
	\end{tabular}
	\caption{Performance metrics with Concatenation Fusion on RML dataset}
	\label{tab : Concatenation Fusion on RML}
\end{table}

\paragraph{Weighted Average Feature Fusion}

Given the promising performance of weighted average fusion reported in our earlier work~\cite{ahmad2023multilevel} for stress classification, this fusion approach was further investigated in the current study. The results with Weighted Average fusion are shown in Table~\ref{tab : Weighted Average Fusion on RML}.
\begin{table}[h]
	\centering
    \begin{tabular}{c c c c c}	
        \hline
        \textbf{Testing Sub} & \textbf{Accuracy (\%)} & \textbf{Precision (\%)} & \textbf{Recall (\%)} & \textbf{$F_1$ Score (\%)} \\
        \hline 
        2 & 75  & 80 & 74.8  & 75.7   \\
        \hline 
        3 & 85.9  & 87.9 & 85.6   & 85.9    \\
        \hline 
        4 & 91.5  & 92.9 & 91.3 &91.2  \\
        \hline 
        5 & 87.2  & 88.2 & 87.1  & 87  \\
        \hline 
        11 &75  & 85.4 & 73.4 & 74.9  \\
        \hline 
        12 & 79.3 & 84.7 & 77.5 & 77.2   \\
        \hline 
        13& 81.25 & 86.6  & 81.2  & 81.8  \\
        \hline 
        14& 79.7 & 87.6  & 79.3  & 80.2  \\
        \hline 
        16 & 79.7  & 85.9  & 79.4  & 80.2\\
        \hline 
        \textbf{Average} & \textbf{81.6} & \textbf{86.6} & \textbf{81.1} & \textbf{81.6} \\
        \hline 
	\end{tabular}
	\caption{Performance metrics with Weighted Average Fusion on RML dataset}
	\label{tab : Weighted Average Fusion on RML}
\end{table}

\paragraph{Gated Average Feature Fusion}

Building upon the effectiveness of Gated Average Fusion (GAF) demonstrated in~\cite{ahmad2020cnn}, we employed this fusion technique to combine the morphological features extracted by the 1D-CNN with the temporal features learned by the BiLSTM-Attention model.

The results with Gated Average Fusion (GAF) are shown in Table~\ref{tab : Gated Average Fusion on RML}

\begin{table}[h]
	\centering
    \begin{tabular}{c c c c c}	
        \hline
        \textbf{Testing Sub} & \textbf{Accuracy (\%)} & \textbf{Precision (\%)} & \textbf{Recall (\%)} & \textbf{$F_1$ Score (\%)} \\
        \hline 
        2 & 75  & 80.6 & 74.8  & 75.7   \\
        \hline 
        3 & 84.4  & 85.1 & 84.2   & 84.5    \\
        \hline 
        4 & 92.9  & 94.2 & 92.7 &92.8   \\
        \hline 
        5 & 91.5  & 92.5 & 91.4  & 91.5  \\
        \hline 
        11 &67.3  & 71.5 & 67 & 67.5  \\
        \hline 
        12 & 81 & 86.5 & 79.2 & 78.7   \\
        \hline 
        13& 81.2 & 85  & 81.2  & 81.8  \\
        \hline 
        14& 71.9 & 75.7  & 71.7  & 72.8  \\
        \hline 
        16 & 81.2  & 84.5  & 81  & 81.8 \\
        \hline 
        \textbf{Average} & \textbf{80.7} & \textbf{84} & \textbf{80.4} & \textbf{80.8} \\
        \hline 
	\end{tabular}
	\caption{Performance metrics with Gated Average Fusion on RML dataset}
	\label{tab : Gated Average Fusion on RML}
\end{table}

\subsubsection{Evaluation of the Proposed Deep Fusion Frameworks}

We performed expaeriments with the propsed deep learning based fusion framework explained in section~\ref{ sec : Proposed Methods}. The results are shown in Tables~\ref{tab : Residual Fusion on RML},~\ref{tab : Cross Attention Fusion on RML} and~\ref{tab : Residual Cross Attention Fusion on RML}

\begin{table}[h]
	\centering
    \begin{tabular}{c c c c c}	
        \hline
        \textbf{Testing Sub} & \textbf{Accuracy (\%)} & \textbf{Precision (\%)} & \textbf{Recall (\%)} & \textbf{$F_1$ Score (\%)} \\
        \hline 
        2 & 75  & 80.2 & 74.8  & 75.7   \\
        \hline 
        3 & 82.8  & 85.1 & 82.6   & 83    \\
        \hline 
        4 & 93  & 94.2 & 92.7 &92.5   \\
        \hline 
        5 & 90.1  & 90.7 & 90  & 90  \\
        \hline 
        11 &69.2  & 73.3 & 68.8 & 69.2  \\
        \hline 
        12 & 82.7 & 87.4 & 80.8 & 80.1   \\
        \hline 
        13& 82.8 & 87.4  & 82.9  & 83.3  \\
        \hline 
        14& 76.6 & 82.1  & 76.2  & 77.3  \\
        \hline 
        16 & 79.7  & 83.7  & 79.4  & 80.2 \\
        \hline 
        \textbf{Average} & \textbf{81.3} & \textbf{84.9} & \textbf{80.9} & \textbf{81.3} \\
        \hline 
	\end{tabular}
	\caption{Performance metrics with Residual Feature Fusion on RML dataset}
	\label{tab : Residual Fusion on RML}
\end{table}

\begin{table}[h]
	\centering
    \begin{tabular}{c c c c c}	
        \hline
        \textbf{Testing Sub} & \textbf{Accuracy (\%)} & \textbf{Precision (\%)} & \textbf{Recall (\%)} & \textbf{$F_1$ Score (\%)} \\
        \hline 
        2 & 78.1  & 87 & 77.7  & 78.6   \\
        \hline 
        3 & 82.8  & 83.9 & 82.6   & 82.9    \\
        \hline 
        4 & 83.1  & 88.9 & 82.8 &83.3   \\
        \hline 
        5 & 87.3  & 88.6 & 87.3  & 87.1  \\
        \hline 
        11 &75  & 86.4 & 73.4 & 74.9  \\
        \hline 
        12 & 75.9 & 82.8 & 74.3 & 74.3   \\
        \hline 
        13& 79.7 & 85.8  & 79.5  & 80.2  \\
        \hline 
        14& 75 & 82.6  & 74.5  & 75.6  \\
        \hline 
        16 & 79.7  & 87.6  & 79.4  & 80.2 \\
        \hline 
        \textbf{Average} & \textbf{79.6} & \textbf{86} & \textbf{79.1} & \textbf{79.7} \\
        \hline 
		\end{tabular}
	\caption{Performance metrics with BFCAF on RML dataset}
	\label{tab : Cross Attention Fusion on RML}
\end{table}

\begin{table}[h]
	\centering
    \begin{tabular}{c c c c c}	
        \hline
        \textbf{Testing Sub} & \textbf{Accuracy (\%)} & \textbf{Precision (\%)} & \textbf{Recall (\%)} & \textbf{$F_1$ Score (\%)} \\
        \hline 
        2 & 76.6  & 84.2 & 76.2  & 77.1   \\
        \hline 
        3 & 87.6  & 87.8 & 87.6   & 87.7    \\
        \hline 
        4 & 92.9  & 94.9 & 92.7 &92.8   \\
        \hline 
        5 & 90.1  & 90.7 & 90  & 90  \\
        \hline 
        11 &75.1  & 80.3 & 73.7 & 72.1  \\
        \hline 
        12 & 82.7 & 88.1 & 80.8 & 80.3   \\
        \hline 
        13& 82.8 & 85.8  & 82.8  & 82.8  \\
        \hline 
        14& 79.1 & 88  & 78.6  & 79.4  \\
        \hline 
        16 & 81.2  & 85.5  & 80.9  & 81.7 \\
        \hline 
        \textbf{Average} & \textbf{83.1} & \textbf{87.2} & \textbf{82.6} & \textbf{82.7} \\
        \hline 	
		\end{tabular}
	\caption{Performance metrics with Cross Attention Residual Fusion on RML dataset}
	\label{tab : Residual Cross Attention Fusion on RML}
\end{table}

\subsection{WESAD Dataset}

The WESAD dataset is a multimodal affective computing dataset comprising physiological signals acquired using both chest-worn and wrist-worn sensors. The chest-worn device records electrocardiogram (ECG), accelerometer (ACC), electromyogram (EMG), respiration, and body temperature signals, whereas the wrist-worn device captures photoplethysmogram (PPG), accelerometer, electrodermal activity (EDA), and body temperature measurements. Data were collected from 17 participants, each undergoing a protocol of approximately two hours. However, due to sensor malfunction, recordings from Subjects 1 and 12 were excluded from the dataset. In this study, only the raw ECG signals obtained from the chest-worn sensor were utilized. Experiments were conducted to evaluate the proposed framework on three class stress classification. 

For the WESAD dataset, experiments were conducted using the 1D-DCNN, BiLSTM, and Residual Cross-Attention Fusion Framework only. This decision was motivated by the superior performance of the Residual Cross-Attention Fusion Framework among all proposed deep feature fusion approaches when evaluated on the RML dataset. As demonstrated in Table~\ref{tab : Residual Cross Attention Fusion on RML}, the Residual Cross-Attention Fusion Framework consistently outperformed the alternative fusion strategies and was therefore selected for further evaluation on the WESAD dataset.

The experimental results obtained using the 1D-DCNN, BiLSTM, and Residual Cross-Attention Fusion Framework are presented in Tables~\ref{tab:1DCNN on WESAD},~\ref{tab: LSTM on WESAD}. and~\ref{tab: Residual Cross Attention Fusion on WESAD}

\begin{table}[h]
\centering
    \begin{tabular}{c c c c c} \hline 
        \textbf{Testing Sub} & \textbf{Accuracy (\%)} & \textbf{Precision (\%)} & \textbf{Recall (\%)} & \textbf{$F_1$ Score (\%)} \\
        \hline  
        2 & 67.30 & 76.03 & 67.30 & 64.50 \\
        \hline  
        3 & 64.52 & 69.61 & 64.52 & 61.60 \\
        \hline  
        4 & 59.18 & 61.97 & 59.18 & 56.37 \\
        \hline  
        5 & 68.28 & 65.98 & 68.28 & 65.17 \\
        \hline  
        6 & 53.93 & 50.44 & 53.93 & 47.09 \\
        \hline  
        7 & 63.12 & 59.39 & 63.12 & 57.06 \\
        \hline  
        8 & 75.73 & 75.53 & 75.73 & 72.81 \\
        \hline  
        9 & 46.63 & 46.80 & 46.63 & 45.00 \\
        \hline  
        10 & 51.54 & 50.49 & 51.54 & 47.91 \\
        \hline  
        11 & 74.24 & 72.11 & 74.24 & 70.75 \\
        \hline  
        13 & 58.41 & 54.71 & 58.41 & 51.33 \\
        \hline  
        14 & 74.64 & 72.02 & 74.64 & 68.44 \\
        \hline  
        15 & 59.43 & 58.58 & 59.43 & 56.07 \\
        \hline  
        16 & 74.92 & 74.19 & 74.92 & 66.19 \\
        \hline  
        17 & 78.65 & 79.58 & 78.65 & 76.71 \\
        \hline  
        \textbf{Average} & \textbf{64.71} & \textbf{64.49} & \textbf{64.71} & \textbf{60.46} \\
        \hline  
    \end{tabular}
    \caption{Performance metrics With 1DCNN on WESAD Dataset.} 
    \label{tab:1DCNN on WESAD} 
\end{table}

\begin{table}[h] 
\centering
    \begin{tabular}{c c c c c} 
    \hline \textbf{Testing Sub} & \textbf{Accuracy (\%)} & \textbf{Precision (\%)} & \textbf{Recall (\%)} & \textbf{$F_1$ Score (\%)} \\
        \hline  
        2 & 46.11 & 40.31 & 46.11 & 36.62 \\
        \hline  
        3 & 59.71 & 61.28 & 59.71 & 59.74 \\
        \hline  
        4 & 53.87 & 61.84 & 53.87 & 52.33 \\
        \hline  
        5 & 66.17 & 68.71 & 66.17 & 66.40 \\
        \hline  
        6 & 50.65 & 52.48 & 50.65 & 49.71 \\
        \hline  
        7 & 73.24 & 74.49 & 73.24 & 71.95 \\
        \hline  
        8 & 65.65 & 63.84 & 65.65 & 61.54 \\
        \hline  
        9 & 48.06 & 48.90 & 48.06 & 46.74 \\
        \hline  
        10 & 50.23 & 50.84 & 50.23 & 44.31 \\
        \hline  
        11 & 75.61 & 74.24 & 75.61 & 72.25 \\
        \hline  
        13 & 57.27 & 57.56 & 57.27 & 52.89 \\
        \hline  
        14 & 70.79 & 66.79 & 70.79 & 63.46 \\
        \hline  
        15 & 49.61 & 47.25 & 49.61 & 46.87 \\
        \hline  
        16 & 74.98 & 73.25 & 74.98 & 68.97 \\
        \hline  
        17 & 76.93 & 73.12 & 76.93 & 72.50 \\
        \hline  
        \textbf{Average} & \textbf{61.26} & \textbf{60.99} & \textbf{61.26} & \textbf{57.75} \\
        \hline  
    \end{tabular}
    \caption{Performance metrics With BiLSTM on WESAD dataset.} 
    \label{tab: LSTM on WESAD} 
\end{table}

\begin{table}[h] 
\centering
    \begin{tabular}{c c c c c} \hline
        \textbf{Testing Sub} & \textbf{Accuracy (\%)} & \textbf{Precision (\%)} & \textbf{Recall (\%)} & \textbf{$F_1$ Score (\%)} \\
        \hline  
        2 & 77.42 & 79.74 & 77.40 & 77.30 \\
        \hline  
        3 & 76.69 & 80.04 & 76.69 & 77.00 \\
        \hline  
        4 & 78.36 & 80.36 & 78.36 & 78.62 \\
        \hline  
        5 & 85.15 & 85.83 & 85.15 & 85.35 \\
        \hline  
        6 & 73.85 & 74.56 & 73.85 & 73.24 \\
        \hline  
        7 & 74.25 & 77.10 & 74.25 & 75.01 \\
        \hline  
        8 & 74.46 & 77.30 & 74.46 & 74.25 \\
        \hline  
        9 & 70.90 & 73.85 & 70.90 & 71.51 \\
        \hline  
        10 & 67.65 & 69.89 & 67.65 & 67.65 \\
        \hline  
        11 & 84.82 & 86.04 & 84.82 & 84.61 \\
        \hline  
        13 & 76.89 & 80.26 & 76.89 & 76.69 \\
        \hline  
        14 & 84.00 & 85.32 & 84.00 & 83.90 \\
        \hline  
        15 & 71.51 & 73.03 & 71.51 & 71.41 \\
        \hline  
        16 & 85.32 & 88.68 & 85.32 & 85.22 \\
        \hline  
        17 & 87.25 & 88.07 & 87.25 & 87.05 \\
        \hline  
        \textbf{Average} & \textbf{77.91} & \textbf{80.04} & \textbf{77.91} & \textbf{78.01} \\
        \hline  
    \end{tabular}
    \caption{Performance metrics With Cross Attention Residual Fusion on WESAD Dataset.}
    \label{tab: Residual Cross Attention Fusion on WESAD}
\end{table}

\section{Discussion} 
\subsection{Quantitative Analysis} 

The results demonstrate the effectiveness of the proposed feature fusion approaches under the subject-independent LOSOCV evaluation protocol. On the RML dataset, CARF achieved the highest average accuracy of 83.1\%, outperforming RFF (81.3\%) and BFCAF (79.6\%). This corresponds to improvements of 1.8 and 3.5 percentage points over RFF and BFCAF, respectively. These findings suggest that combining residual feature preservation with bidirectional cross-attention is more effective than relying on either fusion strategy independently. An interesting observation from the RML results is that BFCAF performed below RFF. This indicates that introducing cross-attention alone does not necessarily improve classification performance. Preserving discriminative information learned by the individual feature extraction branches appears to play an important role when integrating morphological and temporal ECG representations. The superior performance of CARF suggests that combining feature preservation with adaptive information exchange provides a more effective fusion mechanism. The WESAD results further highlight the advantage of combining the two feature extraction branches. CARF achieved an average accuracy of 77.91\%, compared with 64.71\% for the CNN branch and 61.26\% for the BiLSTM-Attention branch. These results represent performance gains of 13.20 and 16.65 percentage points, respectively, demonstrating that the fused representation captures more discriminative stress-related information than either branch alone. A comparison across the two datasets reveals differences in classification performance. While CARF achieved an average accuracy of 83.1\% on the RML dataset, its accuracy on WESAD was 77.91\%. These variations are likely influenced by differences in dataset characteristics, including data distributions, acquisition conditions, and subject-level variability. Such factors can affect the ability of learned representations to distinguish stress levels across previously unseen individuals. Nevertheless, CARF consistently achieved the best overall performance, outperforming the competing fusion strategies on RML and substantially exceeding the individual feature extraction branches on WESAD. Overall, the results demonstrate that integrating residual feature preservation with bidirectional cross-attention provides an effective and robust framework for subject-independent ECG-based stress classification.

 \subsection{Qualitative Analysis} 
 
The results highlight the value of combining morphological and temporal information for ECG-based stress classification. The 1D-CNN branch learns local patterns from ECG waveforms, including morphological characteristics, while the BiLSTM-Attention branch captures temporal relationships and contextual information across the signal. These two representations provide complementary views of the ECG signal and may help the classification model distinguish stress-related patterns more effectively than relying on either representation alone.

The differences in performance among the three fusion frameworks also provide useful insight into how these representations should be combined. On the RML dataset, RFF performed better than BFCAF, suggesting that feature interaction through cross-attention alone may not always provide sufficient benefit. Preserving the discriminative information learned by the individual branches appears to be an important consideration during feature integration. CARF addresses both aspects by retaining the original branch information through residual connections while allowing the two feature representations to exchange information through bidirectional cross-attention. Its higher performance suggests that combining information preservation with adaptive feature interaction can provide a more effective representation for stress classification.

The WESAD results provide further support for this observation. CARF achieved higher classification accuracy than either the CNN or BiLSTM-Attention branch across the evaluated subjects. This indicates that the benefit of fusion is not limited to the RML dataset and that combining morphological and temporal representations can improve classification performance under different subject-independent evaluation settings. However, the differences in performance between individual subjects also highlight the challenges associated with subject-independent stress classification. Variations in ECG characteristics and subject-level signal patterns may affect how consistently a model identifies stress-related information across individuals.

Overall, the findings suggest that the effectiveness of feature fusion depends not only on combining different representations but also on how information is preserved and exchanged between them. CARF provides a unified approach to both objectives, which may explain its stronger performance on the two datasets. Although further investigation is needed to establish how the learned features relate to specific physiological changes, the results support the use of residual learning and bidirectional cross-attention as complementary mechanisms for subject-independent ECG stress classification.

\begin{table}[h]
	\centering
	\begin{tabular}{c c c}
		\hline
		\textbf{Previous Methods} & \textbf{Accuracy (\%)}  & \textbf{$F_1$ Score(\%)} \\
        \hline 
		~\cite{ahmad2023multilevel}     &      66.6 & 65.6\\
        \hline 
		~\cite{ahmad2025stress}     &      71.01 & 70.5\\
        \hline 
		\multicolumn{3}{l}{\textbf{\textit{Proposed Fusion Frameworks}}}\\
		\hline
		BFCAF Framework & 79.6 & 79.7 \\
        \hline 
		Residual Fusion Framework  & 81.3 & 81.3 \\
        \hline 
		\makecell{\textbf{Cross Attention Residual} \\ \textbf{ Fusion Framework}}   & \textbf{83.1} &  \textbf{82.7} \\
        \hline 			
	\end{tabular}
	\caption{Performance comparison of the proposed method On the RML dataset with previous methods for stress classification using LOSOCV splitting.}
	\label{tab:comparison RML Dataset}
\end{table}

\begin{table}[!htbp]
	\centering
	\setlength{\tabcolsep}{3pt}
	\begin{tabular}{c c c}
        \hline
		\textbf{Previous Methods} & \textbf{Accuracy (\%)} & \textbf{$F_1$ Score (\%)} \\
        \hline  
		~\cite{schmidt2018introducing}   &  66.29 & 56.03  \\
        \hline 
		~\cite{garg2021stress}   &  67.56 & 65.73  \\
        \hline 
		~\cite{ahmad2023multilevel}  &  72.7 & 73.1  \\
        \hline 
		~\cite{ahmad2025stress}  &  76.7 & 76.8  \\
        \hline 
        \makecell{\textbf{Cross Attention Residual} \\ \textbf{ Fusion Framework}}  & \textbf{77.91} & \textbf{78.01} \\
        \hline 	 		
	\end{tabular}
	\caption{Performance comparison of the proposed method On the WESAD dataset with previous methods for three class stress classification using LOSOCV splitting.}
	\label{tab:comparison 3class wesad}
\end{table}

\section{Conclusion}

In this paper, we presented three deep learning-based feature fusion frameworks, namely Residual Feature Fusion (RFF), Bidirectional Feature-Wise Cross-Attention Fusion (BFCAF), and the proposed Cross-Attention Residual Fusion (CARF). A dual-branch feature extraction framework consisting of a 1D-CNN and a BiLSTM-Attention network was employed to learn complementary morphological and temporal representations from raw ECG signals, respectively. Building upon these heterogeneous representations, the proposed fusion frameworks were designed to enhance feature interaction while preserving discriminative information for stress classification.

Extensive experiments were conducted on the RML and WESAD datasets using a Leave-One-Subject-Out Cross-Validation (LOSOCV) strategy to assess subject-independent generalization performance. The experimental results demonstrated that all proposed fusion strategies consistently outperformed individual feature extraction models, confirming the benefit of integrating complementary ECG representations. Among the investigated approaches, the CARF framework achieved the best overall performance by effectively combining residual learning and bidirectional cross-attention, resulting in improved feature interaction, information preservation, and stress discrimination capability. The obtained results also demonstrated competitive performance compared with existing state-of-the-art methods for three-class stress classification.

The consistent performance improvements observed across both datasets suggest that the proposed fusion strategy captures stress-related physiological patterns that generalize to unseen subjects, highlighting its potential for practical ECG-based stress monitoring applications. In our future work, we are planning to  investigate multimodal fusion, model compression techniques, and cross-dataset generalization to further improve robustness in real-world environments.

\subsubsection*{Broader Impact Statement}

This work investigates feature fusion strategies for improving subject-independent ECG-based stress recognition. More reliable stress monitoring could support applications in healthcare, workplace well-being, and personal health management by helping identify prolonged stress and enabling timely intervention. However, stress prediction models can be affected by individual physiological differences, signal quality, and data collection conditions, and should therefore be used as supportive tools rather than as standalone diagnostic systems. In addition, ECG data may contain sensitive personal health information, making privacy, informed consent, and secure data handling important considerations. Overall, the proposed framework is intended to support beneficial stress-monitoring applications while encouraging the responsible use of physiological data.


\begin{thebibliography}{63}
\providecommand{\natexlab}[1]{#1}
\providecommand{\url}[1]{\texttt{#1}}
\expandafter\ifx\csname urlstyle\endcsname\relax
  \providecommand{\doi}[1]{doi: #1}\else
  \providecommand{\doi}{doi: \begingroup \urlstyle{rm}\Url}\fi

\bibitem[Abidi et~al.(2026)Abidi, Jamaoui, Nouira, and Bedoui]{abidi2026multi}
Afef Abidi, Hosni Jamaoui, Ibtihel Nouira, and Mohamed~H{\'e}di Bedoui.
\newblock Multi-modal stress classification using a dual-stream cnn-lstm network with mid-level fusion and temporal attention.
\newblock In \emph{2026 International Symposium of Systems, Advanced Technologies and Knowledge (ISSATK)}, pp.\  1--6. IEEE, 2026.

\bibitem[Acharya et~al.(2017)Acharya, Oh, Hagiwara, Tan, Adam, Gertych, and San~Tan]{acharya2017deep}
U~Rajendra Acharya, Shu~Lih Oh, Yuki Hagiwara, Jen~Hong Tan, Muhammad Adam, Arkadiusz Gertych, and Ru~San~Tan.
\newblock A deep convolutional neural network model to classify heartbeats.
\newblock \emph{Computers in biology and medicine}, 89:\penalty0 389--396, 2017.

\bibitem[Ahmad \& Khan(2020)Ahmad and Khan]{ahmad2020cnn}
Zeeshan Ahmad and Naimul Khan.
\newblock Cnn-based multistage gated average fusion (mgaf) for human action recognition using depth and inertial sensors.
\newblock \emph{IEEE Sensors Journal}, 21\penalty0 (3):\penalty0 3623--3634, 2020.

\bibitem[Ahmad \& Khan(2022)Ahmad and Khan]{ahmad2022survey}
Zeeshan Ahmad and Naimul Khan.
\newblock A survey on physiological signal-based emotion recognition.
\newblock \emph{Bioengineering}, 9\penalty0 (11):\penalty0 688, 2022.

\bibitem[Ahmad \& Khan(2025)Ahmad and Khan]{ahmad2025stress}
Zeeshan Ahmad and Naimul Khan.
\newblock Stress classification from ecg signals using vision transformer.
\newblock \emph{IEEE Journal of Selected Areas in Sensors}, 3:\penalty0 47--56, 2025.

\bibitem[Ahmad et~al.(2023)Ahmad, Rabbani, Zafar, Ishaque, Krishnan, and Khan]{ahmad2023multilevel}
Zeeshan Ahmad, Suha Rabbani, Muhammad~Rehman Zafar, Syem Ishaque, Sri Krishnan, and Naimul Khan.
\newblock Multilevel stress assessment from ecg in a virtual reality environment using multimodal fusion.
\newblock \emph{IEEE Sensors Journal}, 23\penalty0 (23):\penalty0 29559--29570, 2023.

\bibitem[Akkaya(2026)]{akkaya2026calibration}
Ahmet Akkaya.
\newblock Calibration, not architecture, limits cross-subject wearable stress detection: a multi-dataset, decision-focused evaluation with label-efficient recalibration.
\newblock \emph{BMC Medical Informatics and Decision Making}, 2026.

\bibitem[Bipro et~al.(2026)Bipro, Motin, Santo~Ali, Kabir, Hasanuzzaman, Wang, and Mahmud]{bipro2026attention}
Sapnil~Sarker Bipro, S~Mohammod~Abdul Motin, Md~Santo~Ali, Sumaiya Kabir, GKM Hasanuzzaman, Shuqiang Wang, and Mufti Mahmud.
\newblock An attention driven deep learning architecture for stress monitoring using single channel electrocardiogram signal.
\newblock \emph{IEEE Transactions on Consumer Electronics}, 2026.

\bibitem[Birkenmaier et~al.(2026)Birkenmaier, Kanuganti, and Stork]{birkenmaier2026hybrid}
Dennis Birkenmaier, Shanthan~Rao Kanuganti, and Wilhelm Stork.
\newblock Hybrid feature learning for wearable stress detection: Combining domain knowledge with supervised deep learning.
\newblock \emph{Sensors}, 26\penalty0 (11):\penalty0 3451, 2026.

\bibitem[Cho et~al.(2019)Cho, Park, Dong, and Youn]{cho2019ambulatory}
H.-M. Cho, H.~Park, S.-Y. Dong, and I.~Youn.
\newblock Ambulatory and laboratory stress detection based on raw electrocardiogram signals using a convolutional neural network.
\newblock \emph{Sensors}, 19\penalty0 (20):\penalty0 4408, 2019.

\bibitem[Deng et~al.(2025)Deng, Li, Luo, and Wu]{deng2025efficientecg}
Hanhui Deng, Xinglin Li, Jie Luo, and Di~Wu.
\newblock Efficientecg: Cross-attention with feature fusion for efficient electrocardiogram classification.
\newblock \emph{arXiv preprint arXiv:2512.03804}, 2025.

\bibitem[Ding et~al.(2025)Ding, Ma, and Li]{ding2025multimodal}
Shihang Ding, Lin Ma, and Haifeng Li.
\newblock Multimodal physiological signal emotion recognition based on multi-head cross attention with representation learning.
\newblock \emph{Frontiers in Psychiatry}, 16:\penalty0 1713559, 2025.

\bibitem[Dong \& Xie(2026)Dong and Xie]{dong2026multimodal}
Ruowang Dong and Li~Xie.
\newblock A multimodal cross-attention-based cnn-lstm network for arrhythmia classification and clinical diagnostic support.
\newblock \emph{Biomedical Signal Processing and Control}, 112:\penalty0 108697, 2026.

\bibitem[Garg et~al.(2021)Garg, Santhosh, Dengel, and Ishimaru]{garg2021stress}
Prerna Garg, Jayasankar Santhosh, Andreas Dengel, and Shoya Ishimaru.
\newblock Stress detection by machine learning and wearable sensors.
\newblock In \emph{Companion proceedings of the 26th international conference on intelligent user interfaces}, pp.\  43--45, 2021.

\bibitem[Giakoumis et~al.(2012)Giakoumis, Drosou, Cipresso, Tzovaras, Hassapis, Gaggioli, and Riva]{giakoumis2012using}
D.~Giakoumis, A.~Drosou, P.~Cipresso, D.~Tzovaras, G.~Hassapis, A.~Gaggioli, and G.~Riva.
\newblock Using activity-related behavioural features towards more effective automatic stress detection.
\newblock \emph{PLoS ONE}, 2012.

\bibitem[Giannakakis et~al.(2019)Giannakakis, Trivizakis, Tsiknakis, and Marias]{giannakakis2019novel}
G.~Giannakakis, E.~Trivizakis, M.~Tsiknakis, and K.~Marias.
\newblock A novel multi-kernel {1D} convolutional neural network for stress recognition from {ECG}.
\newblock In \emph{2019 8th International Conference on Affective Computing and Intelligent Interaction Workshops and Demos (ACIIW)}, pp.\  1--4, 2019.

\bibitem[Han et~al.(2024)Han, Ma, Wang, Hou, Li, and Liu]{han2024online}
Shuze Han, Caiyun Ma, Zhongyu Wang, Yutong Hou, Jianqing Li, and Chengyu Liu.
\newblock Online co-learning strategy based on multibranch multiscale network for 12-lead ecg classification.
\newblock \emph{IEEE Transactions on Instrumentation and Measurement}, 73:\penalty0 1--12, 2024.

\bibitem[Hassoon et~al.(2026)Hassoon, Xiong, Hasson, Salahudeen, Abdulsalami, and Khan]{hassoon2026odual}
Basheer~A Hassoon, Shengwu Xiong, Mushtaq~A Hasson, Ridwan Salahudeen, Aminu~Onimisi Abdulsalami, and Tauqeer Khan.
\newblock odual-branch bidirectional attention fusion of temporal and hierarchical representations for robust ecg arrhythmia classification.
\newblock \emph{Expert Systems with Applications}, pp.\  133409, 2026.

\bibitem[He et~al.(2019)He, Li, Liao, Zhang, and Jiang]{he2019real}
J.~He, K.~Li, X.~Liao, P.~Zhang, and N.~Jiang.
\newblock Real-time detection of acute cognitive stress using a convolutional neural network from electrocardiographic signal.
\newblock \emph{IEEE Access}, 7:\penalty0 42710--42717, 2019.

\bibitem[He et~al.(2016)He, Zhang, Ren, and Sun]{he2016deep}
Kaiming He, Xiangyu Zhang, Shaoqing Ren, and Jian Sun.
\newblock Deep residual learning for image recognition.
\newblock In \emph{Proceedings of the IEEE conference on computer vision and pattern recognition}, pp.\  770--778, 2016.

\bibitem[Hochreiter \& Schmidhuber(1997)Hochreiter and Schmidhuber]{hochreiter1997long}
Sepp Hochreiter and J{\"u}rgen Schmidhuber.
\newblock Long short-term memory.
\newblock \emph{Neural computation}, 9\penalty0 (8):\penalty0 1735--1780, 1997.

\bibitem[Huynh et~al.(2021)Huynh, Nguyen, Nguyen, Pirttikangas, and Siirtola]{huynh2021stressnas}
L.~Huynh, T.~Nguyen, T.~Nguyen, S.~Pirttikangas, and P.~Siirtola.
\newblock {StressNAS}: Affect state and stress detection using neural architecture search.
\newblock In \emph{Adjunct Proceedings of the 2021 ACM International Joint Conference on Pervasive and Ubiquitous Computing}, pp.\  121--125, 2021.

\bibitem[Hyung et~al.(2025)Hyung, Kim, Kim, and Im]{hyung2025deepattnet}
Wooseok Hyung, Minsu Kim, Yesung Kim, and Chang-Hwan Im.
\newblock Deepattnet: deep neural network incorporating cross-attention mechanism for subject-independent mental stress detection in passive brain--computer interfaces using bilateral ear-eeg.
\newblock \emph{Frontiers in Human Neuroscience}, 19:\penalty0 1685087, 2025.

\bibitem[Ju et~al.(2026)Ju, Zhang, Hu, Chen, Yan, and Gzhou]{ju2026differential}
Chen~Gshan Ju, Yuhan Zhang, Yingying Hu, Shuwen Chen, Hu~Ishen Yan, and Yan Gzhou.
\newblock A differential channel-assisted cnn-lstm network with attention fusion for automated ecg diagnosis of myocardial infarction.
\newblock In \emph{2026 International Conference on Intelligent Systems, Communications and Computer Networks (ISCCN)}, pp.\  352--355. IEEE, 2026.

\bibitem[Junior et~al.(2026)Junior, U-Vangsy, and Bin]{junior2026multimodal}
Milembolo~Miantezila Junior, Mavoungou Bayone Don-Faustin U-Vangsy, and Guo Bin.
\newblock When multimodal fusion helps: An ablation study of eeg--ecg fusion strategies for emotion recognition.
\newblock \emph{Computers in Biology and Medicine}, 214:\penalty0 111895, 2026.

\bibitem[Kachuee et~al.(2018)Kachuee, Fazeli, and Sarrafzadeh]{kachuee2018ecg}
Mohammad Kachuee, Shayan Fazeli, and Majid Sarrafzadeh.
\newblock Ecg heartbeat classification: A deep transferable representation.
\newblock \emph{arXiv preprint arXiv:1805.00794}, 2018.

\bibitem[Khandelwal \& Sharma(2025)Khandelwal and Sharma]{khandelwal2025machine}
Megha Khandelwal and Arun Sharma.
\newblock Machine learning and deep learning techniques to detect mental stress using various physiological signals: A critical insight.
\newblock \emph{Wiley Interdisciplinary Reviews: Data Mining and Knowledge Discovery}, 15\penalty0 (3):\penalty0 e70035, 2025.

\bibitem[Khorasani \& Demazure(2026)Khorasani and Demazure]{khorasani2026cross}
Arian Khorasani and Th{\'e}ophile Demazure.
\newblock Cross-temporal attention fusion (ctaf) for multimodal physiological signals in self-supervised learning.
\newblock \emph{arXiv preprint arXiv:2602.02784}, 2026.

\bibitem[Kiranyaz et~al.(2015)Kiranyaz, Ince, and Gabbouj]{kiranyaz2015real}
Serkan Kiranyaz, Turker Ince, and Moncef Gabbouj.
\newblock Real-time patient-specific ecg classification by 1-d convolutional neural networks.
\newblock \emph{IEEE transactions on biomedical engineering}, 63\penalty0 (3):\penalty0 664--675, 2015.

\bibitem[K{\i}ym{\i}k(2026)]{kiymik2026dual}
Erkan K{\i}ym{\i}k.
\newblock Dual warm-start fusion versus attention-based fusion in low-label ecg--pcg classification: a controlled ablation study.
\newblock \emph{Physiological Measurement}, 47\penalty0 (7):\penalty0 075010, 2026.

\bibitem[Kotegar et~al.(2026)Kotegar, Bilakeri, et~al.]{kotegar2026cares}
Karunakar~A Kotegar, Shavantrevva Bilakeri, et~al.
\newblock Cares-binet: A multi-scale deep learning framework for ecg arrhythmia classification.
\newblock \emph{Scientific Reports}, 2026.

\bibitem[Kumar \& Ganapathy(2026)Kumar and Ganapathy]{kumar2026trust}
Kaveti~Pavan Kumar and Nagarajan Ganapathy.
\newblock Trust what you fuse: Reliability-aware cross-attention for multimodal physiological stress assessment in the wild.
\newblock In \emph{Proceedings of the IEEE/CVF Conference on Computer Vision and Pattern Recognition}, pp.\  5341--5350, 2026.

\bibitem[Kumari et~al.(2025)Kumari, Ananth, Ramalakshmi, Mohan, Santhoshkumar, and Mahalakshmi]{kumari2025advanced}
L~Krishna Kumari, R~Prem Ananth, K~Ramalakshmi, B~Mohan, SP~Santhoshkumar, and S~Mahalakshmi.
\newblock Advanced epileptic seizure recognition with a hybrid cnn-bilstm model on eeg signals.
\newblock In \emph{2025 IEEE 1st International Conference on Smart and Sustainable Developments in Electrical Engineering (SSDEE)}, pp.\  1--5. IEEE, 2025.

\bibitem[Kurniawan et~al.(2013)Kurniawan, Maslov, and Pechenizkiy]{kurniawan2013stress}
H.~Kurniawan, A.~V. Maslov, and M.~Pechenizkiy.
\newblock Stress detection from speech and galvanic skin response signals.
\newblock In \emph{Proceedings of the 26th IEEE International Symposium on Computer-Based Medical Systems}, pp.\  209--214, 2013.

\bibitem[Kwon et~al.(2026)Kwon, Yoon, Hur, and Kang]{kwon2026wellness}
Nowon Kwon, Seonkyung Yoon, Jungyi Hur, and Hye~Seung Kang.
\newblock When wellness stress detection enters healthcare workflows: An iso-heart-rate, cross-corpus evaluation of wrist wearables.
\newblock In \emph{Healthcare}, volume~14, pp.\  2434. MDPI, 2026.

\bibitem[Lee et~al.(2023)Lee, Kwak, Kim, Won, and Lee]{lee2023seriessleepnet}
Minji Lee, Heon-Gyu Kwak, Hyeong-Jin Kim, Dong-Ok Won, and Seong-Whan Lee.
\newblock Seriessleepnet: An eeg time series model with partial data augmentation for automatic sleep stage scoring.
\newblock \emph{Frontiers in physiology}, 14:\penalty0 1188678, 2023.

\bibitem[Li \& Zhang(2025)Li and Zhang]{li2025multimodal}
Fang Li and Dan Zhang.
\newblock Multimodal physiological signals from wearable sensors for affective computing: A systematic review.
\newblock \emph{Intelligent Sports and Health}, 1\penalty0 (4):\penalty0 210--222, 2025.

\bibitem[McDuff et~al.(2014)McDuff, Gontarek, and Picard]{mcduff2014remote}
D.~McDuff, S.~Gontarek, and R.~Picard.
\newblock Remote measurement of cognitive stress via heart rate variability.
\newblock In \emph{2014 36th Annual International Conference of the IEEE Engineering in Medicine and Biology Society}, pp.\  2957--2960, 2014.

\bibitem[Najia \& Faouzi(2025)Najia and Faouzi]{najia2025enhanced}
Mechichi Najia and Benzarti Faouzi.
\newblock An enhanced hybrid model combining cnn, bilstm, and attention mechanism for ecg segment classification.
\newblock \emph{Biomedical Engineering and Computational Biology}, 16:\penalty0 11795972251341051, 2025.

\bibitem[Pareja~Bernal et~al.(2026)Pareja~Bernal, Ben~Souissi, and Golz]{pareja2026machine}
Luis~Fernando Pareja~Bernal, Souhir Ben~Souissi, and Christoph Golz.
\newblock Machine learning-based detection of workplace stress using wearable and multimodal data: a systematic literature review.
\newblock \emph{Frontiers in Artificial Intelligence}, 9:\penalty0 1837195, 2026.

\bibitem[Peng \& Guo(2026)Peng and Guo]{peng2026ecg}
Guandi Peng and Ying Guo.
\newblock Ecg-based emotion recognition using beat-level complementary feature fusion.
\newblock \emph{Sensors}, 26\penalty0 (19):\penalty0 6169, 2026.

\bibitem[Perumalsamy et~al.(2026)Perumalsamy, Deepthi, Deepa, Hari~Prasad, Praveen~Sundar, and Govindarajan]{perumalsamy2026ai}
Manikandaprabhu Perumalsamy, V~Deepthi, MB~Deepa, D~Hari~Prasad, PV~Praveen~Sundar, and Priya Govindarajan.
\newblock Ai-driven multi-model stress detection using physiological dynamics.
\newblock \emph{Journal on Advances in Signal Processing}, 2026\penalty0 (1):\penalty0 42, 2026.

\bibitem[Qu et~al.(2026)Qu, Zhang, Lu, Wang, and Su]{qu2026maf}
Chang Qu, Xin Zhang, Yansong Lu, Yi~Wang, and Chengzhi Su.
\newblock Maf-net: Multimodal cross-attention-based fusion network for cardiovascular disease classification.
\newblock \emph{Plos one}, 21\penalty0 (4):\penalty0 e0345238, 2026.

\bibitem[Rahul \& Hashmi(2025)Rahul and Hashmi]{rahul2025stress}
Sadasivuni~Sai Rahul and Mohammad~Farukh Hashmi.
\newblock Stress detection using physiological signals with cnn, lstm and attention on wesad dataset.
\newblock In \emph{2025 International Conference on Signal Processing, Computation, Electronics, Power and Telecommunication (IConSCEPT)}, pp.\  1--5. IEEE, 2025.

\bibitem[Sahu et~al.(2025)Sahu, Gupta, and Lone]{sahu2025anxiety}
Nilesh~Kumar Sahu, Snehil Gupta, and Haroon~R Lone.
\newblock Are anxiety detection models generalizable? a cross-activity and cross-population study using wearables.
\newblock \emph{arXiv preprint arXiv:2504.03695}, 2025.

\bibitem[Sarkar \& Etemad(2020)Sarkar and Etemad]{sarkar2020self}
P.~Sarkar and A.~Etemad.
\newblock Self-supervised {ECG} representation learning for emotion recognition.
\newblock \emph{IEEE Transactions on Affective Computing}, 2020.

\bibitem[Schmidt et~al.(2018)Schmidt, Reiss, Duerichen, Marberger, and Van~Laerhoven]{schmidt2018introducing}
Philip Schmidt, Attila Reiss, Robert Duerichen, Claus Marberger, and Kristof Van~Laerhoven.
\newblock Introducing wesad, a multimodal dataset for wearable stress and affect detection.
\newblock In \emph{Proceedings of the 20th ACM international conference on multimodal interaction}, pp.\  400--408, 2018.

\bibitem[Schuster \& Paliwal(1997)Schuster and Paliwal]{schuster1997bidirectional}
Mike Schuster and Kuldip~K Paliwal.
\newblock Bidirectional recurrent neural networks.
\newblock \emph{IEEE transactions on Signal Processing}, 45\penalty0 (11):\penalty0 2673--2681, 1997.

\bibitem[Seo et~al.(2022)Seo, Kim, Park, and Park]{seo2022deep}
W.~Seo, N.~Kim, C.~Park, and S.-M. Park.
\newblock Deep learning approach for detecting work-related stress using multimodal signals.
\newblock \emph{IEEE Sensors Journal}, 2022.

\bibitem[Shaposhnyk \& Yanushkevich(2026)Shaposhnyk and Yanushkevich]{11463684}
Olha Shaposhnyk and Svetlana Yanushkevich.
\newblock An explainable attention-based lstm fusion network for cognitive load prediction from multimodal physiological signals.
\newblock In \emph{ICASSP 2026 - 2026 IEEE International Conference on Acoustics, Speech and Signal Processing (ICASSP)}, pp.\  22082--22086, 2026.
\newblock \doi{10.1109/ICASSP55912.2026.11463684}.

\bibitem[Sharan et~al.(2025)Sharan, Xiong, and Cui]{sharan2025cardiac}
Roneel~V Sharan, Hao Xiong, and Hui Cui.
\newblock Cardiac abnormality detection from 12-lead ecg using multi-branch cnn and attention-based feature fusion.
\newblock \emph{IEEE Transactions on Instrumentation and Measurement}, 2025.

\bibitem[Sharma et~al.(2026{\natexlab{a}})Sharma, Janakiraman, and Chen]{sharma2026machine}
Shifali Sharma, Aswin~Kumar Janakiraman, and Lujie~Karen Chen.
\newblock Machine learning frameworks for wearable-based stress modeling in naturalistic settings: Scoping review.
\newblock \emph{JMIR mHealth and uHealth}, 14:\penalty0 e76632, 2026{\natexlab{a}}.

\bibitem[Sharma et~al.(2026{\natexlab{b}})Sharma, Panwar, Donelli, and Jain]{sharma2026attention}
Vedprakash Sharma, Arvind Panwar, Massimo Donelli, and Achin Jain.
\newblock An attention-enhanced cnn-bilstm framework with confidence-gated inference for arrhythmia detection using ecg signals.
\newblock \emph{Discover Computing}, 29\penalty0 (1):\penalty0 830, 2026{\natexlab{b}}.

\bibitem[Sosa et~al.(2026)Sosa, Fontecchio, Chrysikou, and Atchison]{sosa2026beyond}
Santiago Sosa, Adam~K Fontecchio, Evangelia~G Chrysikou, and Jennifer~S Atchison.
\newblock Beyond eda: A systematic review of multimodal sympathetic nervous system arousal classification for stress detection.
\newblock \emph{Sensors}, 26\penalty0 (5):\penalty0 1584, 2026.

\bibitem[Swapna \& Janapati(2026)Swapna and Janapati]{swapna2026adaptive}
Padakanti Swapna and Ravichander Janapati.
\newblock Adaptive multi-domain attention network for single-channel eeg classification.
\newblock In \emph{2026 6th International Conference on Soft Computing for Security Applications (ICSCSA)}, pp.\  503--510. IEEE, 2026.

\bibitem[Tsiknakis(2018)]{tsiknakis2018stress}
M.~Tsiknakis.
\newblock Stress detection from speech using spectral slope measurements.
\newblock \emph{Pervasive Computing Paradigms for Mental Health}, 207:\penalty0 41, 2018.

\bibitem[Vaitheeshwari et~al.(2022)Vaitheeshwari, Yeh, Wu, Chen, and Chung]{vaitheeshwari2022stress}
R.~Vaitheeshwari, S.-C. Yeh, E.~H.-K. Wu, J.-Y. Chen, and C.-R. Chung.
\newblock Stress recognition based on multiphysiological data in high-pressure driving {VR} scene.
\newblock \emph{IEEE Sensors Journal}, 22\penalty0 (20):\penalty0 19897--19907, 2022.

\bibitem[Vaswani et~al.(2017)Vaswani, Shazeer, Parmar, Uszkoreit, Jones, Gomez, Kaiser, and Polosukhin]{vaswani2017attention}
Ashish Vaswani, Noam Shazeer, Niki Parmar, Jakob Uszkoreit, Llion Jones, Aidan~N Gomez, {\L}ukasz Kaiser, and Illia Polosukhin.
\newblock Attention is all you need.
\newblock \emph{Advances in neural information processing systems}, 30, 2017.

\bibitem[Womack \& Hansen(1996)Womack and Hansen]{womack1996classification}
B.~D. Womack and J.~H. Hansen.
\newblock Classification of speech under stress using target driven features.
\newblock \emph{Speech Communication}, 20\penalty0 (1--2):\penalty0 131--150, 1996.

\bibitem[Wu et~al.(2026)Wu, Tong, and Qi]{wu2026novel}
Yuxuan Wu, Jijun Tong, and Pengjia Qi.
\newblock A novel adaptive cnn-lstm fusion network for electrocardiogram diagnosis.
\newblock \emph{Physiological Measurement}, 47\penalty0 (1):\penalty0 015004, 2026.

\bibitem[Xu et~al.(2026)Xu, Song, Yue, and Ding]{xu2026variational}
Juncong Xu, Cheng Song, Zijie Yue, and Shuai Ding.
\newblock Variational instance-adaptive personalized stress recognition based on wearable sensor signals.
\newblock \emph{IEEE Transactions on Biomedical Engineering}, 2026.

\bibitem[Yang et~al.(2025)Yang, Yu, and Sano]{yang2025contrastive}
Zeyu Yang, Han Yu, and Akane Sano.
\newblock Contrastive pretraining for stress detection with multimodal wearable sensor data and surveys.
\newblock \emph{Proceedings of machine learning research}, 287:\penalty0 166, 2025.

\bibitem[Zhao et~al.(2026)Zhao, Wang, Gao, and Wang]{zhao2026grf}
Yanyan Zhao, Danli Wang, Xuange Gao, and Xinyuan Wang.
\newblock Grf-mmps: An emotion-driven decision network based on graph residual fusion using multimodal physiological signals.
\newblock \emph{Knowledge-Based Systems}, pp.\  115264, 2026.

\end{thebibliography}
\end{document}